\documentclass[twoside,11pt]{article}

\usepackage[abbrvbib, preprint]{jmlr2e}

\usepackage{jmlr2e}
\usepackage{amsmath}
\usepackage{graphicx}
\usepackage{amssymb}
\usepackage{amsfonts}
\usepackage{tabularx}

\usepackage{lastpage}
\jmlrheading{23}{2022}{1-\pageref{LastPage}}{1/21; Revised 5/22}{9/22}{21-0000}{Sriram Nagaraj and Truman Hickok}

\ShortHeadings{BrowNNe: Brownian Nonlocal Neurons \& Activation Functions}{Sriram Nagaraj and Truman Hickok}
\firstpageno{1}

\begin{document}

\title{BrowNNe: Brownian Nonlocal Neurons \& Activation Functions}

\author{\name Sriram Nagaraj \email snagaraj@swri.org \\
       \addr Southwest Research Institute\\
       6220 Culebra Road\\
       San Antonio, TX 78238, USA
       \AND
       \name Truman Hickok \email thickok@swri.org \\
       \addr Southwest Research Institute\\
       6220 Culebra Road\\
       San Antonio, TX 78238, USA}

\editor{}

\maketitle

\begin{abstract}
It is generally thought that the use of stochastic activation functions in deep learning architectures yield models with superior generalization abilities. However, a sufficiently rigorous statement and theoretical proof of this heuristic is lacking in the literature. In this paper, we provide several novel contributions to the literature in this regard. Defining a new notion of nonlocal directional derivative, we analyze its theoretical properties (existence and convergence). Second, using a probabilistic reformulation, we show that nonlocal derivatives are epsilon-sub gradients, and derive sample complexity results for convergence of stochastic gradient descent-like methods using nonlocal derivatives. Finally, using our analysis of the nonlocal gradient of H\"older continuous functions, we observe that sample paths of Brownian motion admit nonlocal directional derivatives, and the nonlocal derivatives of Brownian motion are seen to be Gaussian processes with computable mean and standard deviation. Using the theory of nonlocal directional derivatives, we solve a highly nondifferentiable and nonconvex model problem of parameter estimation on image articulation manifolds. Using Brownian motion infused ReLU activation functions with the nonlocal gradient in place of the usual gradient during backpropagation, we also perform experiments on multiple well-studied deep learning architectures. Our experiments indicate the superior generalization capabilities of Brownian neural activation functions in low-training data regimes, where the use of stochastic neurons  beats the deterministic ReLU counterpart. 
\end{abstract}


\section{Introduction}\label{sec:intro}
The success of modern deep learning in solving a plethora of problems is truly stunning. Deep neural network (NN) architectures have been customized and adapted to a multitude of application areas. Indeed, deep learning based analysis has emerged as a third pillar of science along with theory, and experiment. Much of the success of deep learning can be attributed to i) the powerful function approximation properties delivered by NNs as illustrated in \cite{cybenko1989approximation}, \cite{hornik1989multilayer}, \cite{yarotsky2017error} and \cite{m} among other works, ii) the superior generalization capabilities of NN based models as described in \cite{goodfellow}, \cite{shalev2014understanding} and \cite{vidyasagar2002learning} iii) the development of well designed software frameworks that allow for easy training and deployment of deep NN models utilizing significant computing resources (see for e.g. \citealt{bigsurvey}, \citealt{chen1} and \citealt{survey1}) and iv) the emergence of auto-differentiation functionality in modern deep learning software (see \citealt{baydin2018automatic} for a good overview of this topic). At the heart of points i) and ii) above is the use of nonlinear activation functions that define the input-output relationship of a neuron. The flexibility and adaptivity of nonlinear activation functions result in universal function (and, as the recent work \citealt{raissi2019physics} shows, operator) approximations. Training deep NNs is typically done using first-order (gradient of pseudo-gradient based) auto-differentiation (either forward or reverse mode) implemented with computational graphs being the fundamental data architecture. This requires the efficient evaluation of both the nonlinear activation function, and its gradient/pseudo-gradient. 

The emergence of this paradigm has not been without challenges. Considering specifically the auto-differentiation part, the flow of gradient information through the network (forward and back) has shown to be highly sensitive to numerical perturbations. Indeed, the so-called vanishing (and exploding) gradient problem plagued the class of recurrent NN, until the analysis of appropriate initialization conditions. \cite{hochreiter1998vanishing} discuss the vanishing gradient problem while \cite{glorot} considers efficient initialization techniques. Likewise, different classes of nonlinearities used in activation functions yielded vastly differing results in practice, and much work has been done in studying the specific form of ``good" activation functions. \cite{ramachandran2017searching} and \cite{dubey} compare and benchmark several classes of activation functions. For instance, the (nondifferentiable) rectified linear unit (aka ReLu) activation function and its cousins (SiLu, leaky ReLU etc.) have largely displaced the smooth sigmoid nonlinearity. A thorough analysis of nondifferentiable activation functions, both theoretical soundness and practical applicability in the context of automatic differentiation can be found in \cite{pap}. 

\subsection{Irregular activation functions, biased/pseudo gradients and nonlocal calculus}
Results in the literature also point to the promise of using stochastic activation functions, especially for superior generalization capabilities. For example, the results of \cite{bengio} and \cite{probact} consider a class of probabilistic (i.e. noisy) activation functions, and the results show that such activation functions are produce tractable alternatives to standard deterministic ones, while delivering superior generalization. \cite{bengio} considered noisy activation functions, which were shown to reduce the saturation of traditional activation functions similar to simulated annealing. \cite{bengio} considered the question of backpropagating stochastic (non-differentiable) neurons. \cite{probact} looked at probabilistic activation functions, where an activation function's output is obtained from sampling from a distribution over the mean and variance. \cite{probact} also considers the question of how backpropagation can be adapted to such a scenario, and showed the computational benefit of the approach on some benchmark deep learning models. In a similar fashion, \cite{probfn} consider probabilistic NN for image recognition. In \cite{qnn}, the authors consider the positive effects of quantization on NN performance, in particular, stochastic binarization. The use of adaptive activation functions as considered in \cite{adaptivefn} result in increased accuracy in the case of Physics Informed Neural Networks (PINNs). More generally, in \cite{kan}, a new class of NNs have been proposed which depend on the Kolmogorov-Arnold approximation theorem. Indeed, these so-called KANs are trained to learn activation functions themselves, and are not restricted to a fixed form of activation.

The use of subgradients/biased-gradients/pseudo-gradients etc. in place of ``usual" gradients has also been well studied, and these modified gradient descent-like methods underpin much of modern deep learning (see for e.g. \citealt{biased1} and \citealt{biased2}). Roughly, for a reasonable class of optimization problems, given access to ``gradient-like" samples $h_k$ at each iteration $k$, updates of the form $x_{k+1} = x_{k} - \gamma_k h_k$ yield decent convergence results. Often, the samples $h_k$ are required to be bounded and obey relations such as $\mathbb{E}(h_k) = g_k$ where $g_k$ is a true (sub)-gradient.  In addition, pseudo-gradients are also seen to guide gradient-free learning models as well. For instance \cite{niru} considers evolutionary strategies for optimization which are augmented with gradient-like surrogates for increased efficiency.

\subsubsection{Nonlocal operators}

Nonlocal operators encapsulate the notion of action at a distance, and account for the behavior of a system both at and \emph{around} a given point. Specifically, the value of a nonlocal operator $N:X\rightarrow Y$ at $x\in X$ would depend both on $x$ and values in some neighborhood of $x$. Modeling a phenomenon via nonlocal operators has a variety of use cases. Some salient examples include peridynamics (\citealt{alali2012multiscale} and \citealt{emmrich2013peridynamics}), computational mechanics (\citealt{du2019nonlocal},\citealt{bavzant2002nonlocal},\citealt{duddu2013nonlocal}) , convolution, smoothing and filtering in signal/image processing (\citealt{gilboa2009nonlocal}, \citealt{buades2005non}), deep learning (\citealt{tao2018nonlocal}) etc. 
The relationship between nonlocal and fractional models is close. The work of \cite{d2021towards} shows that, in many cases, the two are essentially isomorphic modeling choices.

In our work, we consider nonlocal analogues of differential operators, specifically the directional derivative. \cite{alali2015generalized} consider nonlocal gradient/divergence operators from a vector calculus perspective. \cite{Mengesha2015LocalizationON}, \cite{bellido} and \cite{fods} (among others) use a sequence of radial kernels to define Frechet-like nonlocal differential operators and study the convergence (both the mode and topology) of these operators to their local counterpart in settings where the ``usual" operators exist.

\subsection{Resume of Contributions and Paper Organization}
This paper puts forth a new approach for dealing with nondifferentiable optimization problems, which is applicable in highly irregular situations. At an extreme, the methods developed here are applicable to \emph{nowhere} differentiable objective functions. Indeed, ordinary calculus, including traditional sub/biased/pseudo gradient approaches do not carry over to this case. However, since our analysis relies on the notion of \emph{nonlocal} calculus, we are able to use nonlocal analogs of differential operators that are defined for highly nondifferentiable functions. In the first part of the paper, it is shown that when a ``usual" derivative exists (either in a strong or weak sense), the corresponding (Gateaux) nonlocal analog approximates the standard directional derivative. We define the Gateaux-like nonlocal directional derivative (NDD) of a function $u:\Omega \subset \mathbb{R}^n\rightarrow \mathbb{R}$ along a direction $v\in \mathbb{R}^n$ as \begin{equation}D_{v,\rho}u(x) = \int \frac{u(x+tv)-u(x)}{t}\rho(t)dt,\end{equation}where the kernel $\rho(\cdot)$ is appropriately chosen, and controls the extent of nonlocal interactions around $x$. Indeed, when $\rho(t)=\delta(t),$ the Dirac mass, we recover, in the sense of distributions, the ``usual" directional derivative (assuming the necessary regularity on $u$). The first section of the paper (Section \ref{sec:theory}) analyzes the NDD in detail. Conditions on the existence of the NDD, the mode of convergence of the NDD to the usual directional derivative (when the latter exists), and a nonlocal Taylor approximation are also studied. Convergence is proved for functions in $C^1(\Omega), W^{1,p}(\Omega)$ and Lipschitz continuous functions as well. In addition, existence of NDD for H\"older continuous functions is also proved. It is possible to extend this definition of the nonlocal Gateaux operator to the infinite dimensional case as well, though this would use the Bochner (or the weaker Gelfand-Pettis) integral. Note that such an extension of a ``Frechet" nonlocal gradient (such those defined in \citealt{alali2015generalized}, \citealt{Mengesha2015LocalizationON} and \citealt{bellido}) to infinite dimensions would be more challenging. Indeed, in the absence of an infinite dimensional analogue of Lebesgue measure, this would require the design of appropriate radially symmetric Gaussian measures on Banach/Hilbert spaces. Finally, keeping practical uses of nonlocal derivatives in mind, reducing the computational burden from a high dimensional integral to a one dimensional integral allows for the rapid evaluation of the NDD as presented here. Indeed, as we shall see, the evaluation of NDDs does not require the evaluation of an integral, since they can be computed by sampling techniques. This is yet another (albeit computational) advantage over the nonlocal Frechet-type operators.

In Section \ref{sec:bgd}, we consider a probabilistic reformulation of the NDD:
\begin{equation}D_{v,\rho}u(x) = \mathbb{E}_{\rho}\mathcal{Q}(u,v,x,T),\end{equation}
where $\mathcal{Q}(u,v,x,T) = \frac{u(x+Tv)-u(x)}{T}$, and show how it is a type of biased subgradient. Applications of the NDD in the optimization of non-differentiable objective functions readily follow. Indeed, using this probabilistic reformulation, we next show that our nonlocal gradient is an $\epsilon$-subgradient, and we show that such $\epsilon$-subgradients can be used to perform reasonable first-order descent algorithms. Finally, in the last part of the paper, we consider the specific example of Brownian motion. Section \ref{sec:brownian} is devoted to the study of the NDD of the sample paths of Brownian motion. Using a specific interaction kernel, we show that this NDD of Brownian sample paths exists, and is a zero mean Gaussian process with strictly increasing variance. The basis of this section relies on the existence of the nonlocal derivative for H\"older continuous functions (such as the sample path of Brownian motion). Section \ref{sec:experiments} described several experiments involving 1st order stochastic optimization algorithms, and also standard deep learning architectures and models. We consider the case of image articulation manifolds (IAMs), which are known to be nondifferentiable manifolds of images obtained by imaging a scene under changing imaging conditions. Using the NDD in place of the standard gradient, we first show the ability to perform nondifferentiable (and nonconvex) parameter estimation. Next, we consider the case of standard deep learning architectures where backpropagation is performed with stochastic neurons whose activation function is given by a ReLU + (scaled) Brownian sample path, i.e. of the form $\phi(x)+\alpha W(x)$ where $\phi(\cdot)$ is the usual ReLU unit and $W(x)$ is a zero mean Wiener process, with $\alpha$ being a scaling constant. We show that in low training data regimes ($\leq 50\%$ training data), the use of such activation functions is justified by a increase in generalization performance. We show empirically where Brownian (aka ``nondiff") neurons should be placed in specific architectures to yield appreciable performance gains. Proofs for theorems/propositions/lemmas will be provided in Section \ref{sec:proofs}. 

\section{Nonlocal Theory}\label{sec:theory}
We begin with a review of the notation used in this work, and proceed to establishing the main theoretical devices needed for the rest of the paper.
\subsection{Notation and Review}
We fix a simply connected domain $\Omega \subset \mathbb{R}^{N}$ with a compact, Lipschitz boundary. Thus, $\Omega$ is a simply connected open subset of $\mathbb{R}^{N}$ with boundary $\partial \Omega$ which is compact and can be realized locally as the graph of a Lipschitz continuous function. The dimension $N$ will always be fixed in our discussion with $\mathbb{R}^N$ having coordinates $x_1,\ldots,x_N$, and $\Omega$ is given the subspace topology being in $\mathbb{R}^{N}$. If not otherwise specified, $\|\cdot\|$ will denote the Euclidean norm. Vectors will always be column vectors, and elements of vectors (resp. tensors) will be indexed by subscripts, so that $x_i$ is the $i^{\text{th}}$ element of vector $x$. The transpose of vectors (resp tensors) will be denoted by superscript $T$. The abbreviation ``a.e." is ``almost everywhere" with respect to the Lebesgue measure. All integrals will be in the Lebesgue sense except briefly in Section \ref{sec:nldd}.

For a function $u:\Omega \rightarrow \mathbb{R}$, the notation $\text{supp}(u)$ will denote the support of $u$, i.e., 

\begin{equation}\text{supp}(u) = \overline{\{x\in \Omega: |u(x)|>0 \,\text{a.e.}\}},\end{equation}
and the overbar $\overline{\mathcal{S}}$ of a subset $\mathcal{S}\subset\mathbb{R}^{N}$ is the closure of $\mathcal{S}$ in the Euclidean $\mathbb{R}^{N}$ topology. The essential supremum of $u$ is denoted as $\overline{\text{sup}}(u).$ For a set $X$, we denote by $\mathbb{I}_X$ the indicator function on $X$: \begin{equation*}\mathbb{I}_X(x) = 1, \text { if } x\in X, \, \mathbb{I}_X(x) = 0 \text{ otherwise}.\end{equation*}

Unless otherwise specified, the letters $i,j,k,n,m,N$ will be non-negative integers while $\lambda,\epsilon,\delta,M,K$ will denote arbitrary real values. A multi-index $\overline{n}$ is an ordered tuple of non-negative integers, $\overline{n}=(n_1,\ldots,n_N),$ and we define its magnitude as $|\overline{n}| = \sum_{i=1}^{N}n_i.$

\subsubsection{Function Spaces}\label{sec:fsops}
Given a function $u:\Omega \rightarrow \mathbb{R}$, and a multi-index $\overline{n}=(n_1,\ldots,n_N)$, we denote by $\frac{\partial ^{\overline{n}}u}{\partial x^{\overline{n}}}$ to be: $$ \frac{\partial ^{\overline{n}}u}{\partial x^{\overline{n}}} = \frac{\partial ^{n_1}}{\partial x_{1}^{n_1}}\frac{\partial ^{n_2}}{\partial x_{2}^{n_2}}\ldots \frac{\partial ^{n_N}u}{\partial x_{N}^{n_N}}.$$ In particular, the gradient $\nabla u$ of $u$ corresponds to the collection $\frac{\partial ^{\overline{n_i}}u}{\partial x^{\overline{n_i}}}$ in the case where $\overline{n_i} = \underbrace{(0,0,\ldots,1,0,\ldots,0)}_{1 \text{ in the $i^{\text{th}}$ position}}, \, i = 1,\ldots N$. We will denote the directional derivative of $u$ at $x$ in the direction $v\in \mathbb{R}^N$ as $D_{v}u(x) = v^{T} \nabla u(x).$ For any real $0<  p \leq \infty,$ we define its conjugate $p'$ such that $\frac{1}{p}+ \frac{1}{p'} = 1.$ The function spaces we will encounter are defined as follows:

\begin{equation}
\begin{aligned} 
C^{m}(\Omega) &= \{u:\Omega \rightarrow \mathbb{R}: u \text{ is $m$-times continuously differentiable}\} \\ 
C^{m}_{0}(\Omega) &= \{u \in C^{m}(\Omega):\text{ has compact support inside} \Omega\} \\
L^{p}(\Omega) &= \{u:\Omega \rightarrow \mathbb{R}: \int_{\Omega}|u|^p < \infty\} \\ 
L^{\infty}(\Omega) &= \{u:\Omega \rightarrow \mathbb{R}: \overline{\text{sup}}(u) < \infty\}\\
W^{m,p}(\Omega) &= \{u\in L^{p}(\Omega): \frac{\partial ^{\overline{n}}u}{\partial x^{\overline{n}}} \in L^{p}(\Omega) \, \forall \overline{n}\, , 0\leq |\overline{n}|\leq N \} \\
\text{Lip}(\Omega,M) &= \{u:\Omega \rightarrow \mathbb{R}: u\text{ is Lipschitz continuous with constant }M\} \\
\text{H\"ol}(\Omega,M,\alpha) &= \{u:\Omega \rightarrow \mathbb{R}: u\text{ is H\"older continuous with constant }M \text{ and exponent $\alpha$}\} 
\end{aligned}
\end{equation}
Note that $u\in \text{Lip}(\Omega,M)$ if for all $x,y \in \Omega$ we have that \begin{equation*}{|u(x)-u(y)|<M\|x-y\|}\end{equation*} while $u\in \text{H\"ol}(\Omega,M,\alpha)$ if \begin{equation*}|u(x)-u(y)|<M\|x-y\|^{\alpha}\end{equation*} for all $x,y \in \Omega$. In the sequel, we will be particularly interested in the case where $0 < \alpha < 1.$

As a special case, where $p=2$, we obtain Hilbert spaces $L^{2}(\Omega)$ with inner product $(u,v)_{L^2} = \int_{\Omega}uv$ and $W^{m,2}(\Omega) = H^{m}(\Omega)$ with inner product \begin{equation*}(u,v)_{H^m} = \sum_{|\overline{n}|=0}^{m}(\frac{\partial ^{\overline{n}}u}{\partial x^{\overline{n}}},\frac{\partial ^{\overline{n}}v}{\partial x^{\overline{n}}}),\end{equation*}where the sum is over all multi-indices $\overline{n}$ with a given magnitude. Note that $H^{m}(\Omega)$ can be realized as the closure of $C^{m}(\Omega)$ in the topology generated by $(\cdot,\cdot)_{H^{m}}$. Likewise, the closure of $C^{m}_{0}(\Omega)$ in the same topology gives rise to the subspace $H^{m}_{0}(\Omega) \subset H^{m}(\Omega)$ of functions that ``vanish on the boundary" of $\Omega$. In particular, if $\Omega = \mathbb{R}^N,$ we have that $H^{m}_{0}(\Omega) = H^{m}(\Omega)$. We refer to \cite{adams} for a deep discussion of these ideas. 
\subsubsection{Densities}\label{sec:density}
Our theory of nonlocal directional derivatives relies heavily on the careful choice of a collection $\{\rho_n(\cdot)\}$ of probability densities (weights). As we shall see, each such sequence of densities yields a distinct collection of nonlocal operators. We will work with a collection of densities with appropriate decay conditions on the sequence of tails of the densities, where we define the tail (at location $\delta>0$) of a density $\rho(\cdot)$ as:
\begin{equation}\label{eq:tail}
\text{Tail}_{\delta}(\rho) = \int_{\|x\|>\delta}\rho(x)dx.
\end{equation} Previously (e.g. \citealt{alali2015generalized}, \citealt{Mengesha2015LocalizationON} and \citealt{bellido}, \citealt{fods}), in the nonlocal context, most authors chose to use so-called radial densities to define nonlocal operators. This was in part due to the fact that the nonlocal operators so defined involved a multidimensional integration over $\Omega$ and radial symmetry was needed to prove convergence estimates, while the approach taken here is to use one-dimensional, directional integrals. A radial density is a density in $\mathbb{R}^N$ which is invariant to rotations: $\rho(x) = \rho(Ox)$ for any $N\times N$ orthogonal matrix $O_{N\times N}.$ In other words, this means that $\rho(x) = \psi(\|x\|)$ for some function $\psi:\mathbb{R}\rightarrow \mathbb{R}.$ Thus in one dimension, radial densities are symmetric distribution functions. In the current context, we will not be making this assumption on densities. Indeed, our integrals are one dimensional, and as such do not require radial symmetry. We consider sequences $\{\rho_n\}$ of densities defined on the real line satisfying the following conditions:

\begin{equation}\label{eq:density}
\left\{\begin{aligned} 
&\rho \geq 0,\\
&\int_{\mathbb{R}} \rho(t)dt = 1,\\
&\text{lim}_{n\rightarrow \infty}\text{Tail}_{\delta}(\rho_n) = 0, \, \forall \delta >0.
\end{aligned} \right.
\end{equation}
The last condition in equation \ref{eq:density} requires that the sequence of tails $\text{Tail}_{\delta}(\rho_n)$ vanish as $n\rightarrow \infty$, which, along with the fact that each $\rho_n(\cdot)$ integrate to unity, implies that, informally, the sequence $\{\rho_n\}$ approaches the impulse (or ``Dirac delta") density $\delta_0(t)$ centered at $0$. A great deal of freedom can be exercised in the design of the sequence $\{\rho_n\}$, and we highlight a few examples.
\begin{itemize}
\item Gaussian: $\rho_n(t) = \frac{n}{\sqrt{2\pi}}e^{-\frac{n^2t^2}{2}}.$
\item ``Linear Moving Rectangle": $\rho_{n}(t) = n(n-1)\mathbb{I}_{[\frac{1}{{n}},\frac{1}{{n-1}}]}(t).$
\item ``Exponential Moving Rectangle": $\rho_{n}(t) = 2^{n}\mathbb{I}_{[\frac{1}{2^{n}},\frac{1}{2^{n-1}}]}(t).$

\end{itemize}
Of the three mentioned above, we will make extensive use of the moving rectangle sequences in our analysis of nonlocal directional derivatives, especially as applied to the the Wiener process. Note that both moving rectangle sequences are asymmetric ($\rho_n(t)\neq\rho_n(-t)$, i.e. non-radial in one dimension). This is an important distinction of the nonlocal Gateaux derivative presented here and the prior nonlocal Frechet approaches (e.g. \citealt{alali2015generalized}, \citealt{Mengesha2015LocalizationON} and \citealt{bellido}, \citealt{fods}). Moreover, asymmetric interaction kernels lead to non-trivial results for the NDD of Brownian sample paths, as we shall see shortly.
\begin{proposition}\label{thm:3density}
All three sequences defined above converge in the sense of distributions to the impulse (Dirac delta) density $\delta_0(t)$.
\end{proposition}

\subsection{Nonlocal Directional Derivatives}\label{sec:nldd}
We now begin our study of nonlocal directional derivatives. Given $u:\Omega \rightarrow \mathbb{R}$, we can extend $u$ by zero to all of $\mathbb{R}^N$. This would ensure that $u(x+tv)$ is meaningful for all $t,v$. Moreover, if $u$ is compactly supported in $\Omega$, extension by zero is a canonical extension (\citealt{adams}). Henceforth, we assume $\Omega = \mathbb{R}^N.$ Given an arbitrary density $\rho(\cdot)$ on $\mathbb{R}$, a function $u:\Omega \rightarrow \mathbb{R}, \, x\in \Omega$ and a direction vector $v\in \mathbb{R}^N$, we denote he nonlocal directional derivative of $u$ at $x$ in the direction $v$ as $D_{\rho,v}u(x)$ defined as:
\begin{definition}\label{def:nldd}
\begin{equation}
D_{\rho,v}u(x) = \text{lim}_{\epsilon\rightarrow 0} \int_{\mathbb{R}-B_{\epsilon}(0)}\frac{u(x+tv)-u(x)}{t}\rho(t)dt,
\end{equation}
\end{definition}
where the integral is a Cauchy principal value integral. However, for a large class of functions, the map $t\mapsto \frac{u(x+tv)-u(x)}{t}\rho(t)\in L^{1}(\mathbb{R}),$ and hence the principal value integral agrees with the Lebesgue integral.
\begin{lemma}\label{thm:pvi}
If $u:\Omega \rightarrow \mathbb{R}$ belongs to one of the following spaces: \begin{equation*}C_{0}^1({\Omega}), W^{1,p}(\Omega), \text{Lip}(\Omega,M),\end{equation*} then the principal value integral in Definition \ref{def:nldd} agrees with the Lebesgue integral. In particular, $D_{\rho,v}u(x)$ is bounded. Moreover, if $u\in \text{H\"ol}(\Omega,M,\alpha), \, \alpha>0$ and $\rho(\cdot)$ is such that $\frac{\rho(t)}{|t|^{1-\alpha}}$ is integrable $\mathbb{R}$ (in particular if $\rho(\cdot)$ has compact support), then also, $D_{\rho,v}u(x)$ is bounded.
\end{lemma}
Note that while we have assumed that $u$ is real-valued, Definition \ref{def:nldd} can easily be extended to vector-valued functions, or even functions taking values in infinite dimensional topological vector spaces (TVS). Indeed, we are essentially integrating the map $t\mapsto \frac{u(x+tv)-u(x)}{t}\rho(t)$, and if the integral in Definition \ref{def:nldd} is taken in the sense of the Bochner (or Dunford-Pettis) integrals, this yields a formulation that is valid for functions $u$ with range in vector valued spaces or TVS. In this case, the nonlocal directional derivative in Definition \ref{def:nldd} can be thought of a nonlocal version of the Gateaux derivative. 

Note that we may consider $D_{\rho,v}u(x)$ as acting on $u$, but also parameterized by $v$. As applied on functions $u$, $D_{\rho,v}(\cdot)$ is linear. In general, it is not linear in $v$. Moreover, it is not a derivation, i.e., the product rule does not, in general, exist for $D_{\rho,v}(\cdot)$. Likewise, the chain rule also may not always apply. 

 \begin{lemma}\label{thm:linearity}
The nonlocal directional derivative, $D_{\rho,v}u(x)$ is linear in $u$. Moreover, given $u_1, u_2$, we have that \begin{equation*}D_{\rho,v}(u_1 u_2)(x) = \frac{1}{2}(u_1 D_{\rho,v}(u_2)(x)+u_2 D_{\rho,v}(u_1)(x) + b_{\rho,v}(u_1,u_2)),\end{equation*}where\begin{equation*}b_{\rho,v}(u_1,u_2) \\= \int 2u_1(x+tv)u_2(x+tv)-u_1(x)u_2(x+tv)-u_1(x+tv)u_2(x)\frac{\rho(t)}{t}dt\end{equation*}is bilinear in $(u_1,u_2).$
\end{lemma}

We shall henceforth always assume that $u$ belongs to one of the spaces mentioned in Lemma \ref{thm:pvi} and speak of Lebesgue integrals. While Definition \ref{def:nldd} is valid for any density $\rho(\cdot),$ we will henceforth fix a sequence of densities $\{\rho_n(\cdot)\}$ that satisfy the conditions in \ref{eq:density}, such as one of the families (Gaussian, moving rectangles) mentioned above. The particular family of densities chosen is not very essential for the theoretical development. However, in practice, different families may have widely varying numerical behavior. For now, however, we assume a fixed generic family has been chosen. Having done so, we obtain a sequence of nonlocal directional derivatives, $\{D_{\rho_n,v}u(x)\}.$ To ease notation, we will suppress the $\rho(\cdot)$ and write $D_{n,v}u(x)$ instead of $D_{\rho_n,v}u(x)$, and since the family $\{\rho_n(\cdot)\}$ is fixed, there is no scope for confusion. We now state the main theorem of this section.

\begin{theorem}\label{thm:nlconv}

\begin{itemize}
\item Let $u\in C_{0}^{1}(\Omega)$. Then, $D_{n,v}u\rightarrow D_{v}u$ uniformly.
\item Let $u\in \text{Lip}(\Omega,M)$ with compact support. Then, $D_{n,v}u\rightarrow D_{v}u$ uniformly.
\item Let $u\in W^{1,p}(\Omega)$. Then, $D_{n,v}u\rightarrow D_{v}u$ in $L^p(\Omega)$. 
\end{itemize}

\end{theorem}
 Theorem \ref{thm:nlconv} tells us that, roughly, the nonlocal directional derivative converges, in the appropriate topology, to the standard directional derivative when the latter exists. Moreover, for H\"older continuous functions, which are in general non-differentiable, we can still form the NDD, even though the ``usual" local derivative would not exist. This will be particularly important for our later work involving the Wiener process, whose sample paths are known to be H\"older continuous but nowhere differentiable, and whose subgradients are also not tractable. However, so long as the $D_{n,v}u$ exist, we can choose a large enough $n=n^{*}$ and view $D_{n^{*},v}u$ as a surrogate for the true (non-existent) directional derivative. In this way, we can attempt to perform standard (biased)-gradient based optimization procedures on wildly non-differentiable functions. 
 
 \subsubsection{Nonlocal Taylor's Theorem}\label{sec:nltt}
 We now consider a nonlocal version of Taylor's theorem. Let $v_1,\ldots,v_N \in \mathbb{R}^N.$ We stack the vectors as the rows of matrix $V$:\begin{align}
    V &= \begin{bmatrix}
           -v_{1}^{T}- \\
           -v_{2}^{T}- \\
           \vdots \\
           -v_{N}^{T}-
         \end{bmatrix}
  \end{align}
 Then clearly we have:
 \begin{align}
    V \nabla u (x) &= \begin{bmatrix}
           -D_{v_{1}}u (x)- \\
           -D_{v_{2}}u (x)- \\
           \vdots \\
           -D_{v_{N}}u (x)-
         \end{bmatrix}
  \end{align}
 We define, for a multi-index $\overline{n} = (n_1,\ldots,n_N)$
  \begin{align}
    \nabla_{\overline{n},V}u (x) &= \begin{bmatrix}
           -D_{n_1,v_{1}}u (x)- \\
           -D_{n_2,v_{2}}u (x)- \\
           \vdots \\
           -D_{n_N,v_{N}}u (x)-
         \end{bmatrix}
  \end{align}
From Theorem \ref{thm:nlconv}, we can deduce the following corollary:
\begin{corollary}\label{thm:nlVecconv}

\begin{itemize}
\item Let $u\in C_{0}^{1}(\Omega)$. Then, $\nabla_{\overline{n},V}u\rightarrow V \nabla u $ uniformly as min$(\overline{n})\rightarrow \infty$.
\item Let $u\in \text{Lip}(\Omega,M)$ with compact support. Then, $\nabla_{\overline{n},V}u\rightarrow V \nabla u$ in $L^{\infty}(\Omega)$ almost everywhere as min$(\overline{n})\rightarrow \infty$.

\item Let $u\in W^{1,p}(\Omega)$. Then, $\nabla_{\overline{n},V}u\rightarrow V \nabla u $ in $L^p(\Omega)$ as min$(\overline{n})\rightarrow \infty$. 
\end{itemize}

\end{corollary}
If $V=I_{N\times N}$ is the $N\times N$ identity matrix, we will write $\nabla_{\overline{n},V}u$ as simply $\nabla_{\overline{n}}u$. We can now state the following nonlocal version of Taylor's theorem. 
 
 \begin{theorem}\label{thm:nltaylor}
Let $u\in C^{1}(\Omega)$ and $x_0\in \Omega$. Using a smooth cutoff function $\phi_\eta(\cdot)$ define $\tilde{u}$ such that $\tilde{u}=u$ inside a ball $B_{R}(x_0)$ of radius $R$ around $x_0$ and $\tilde{u} = 0$ outside a compact set $K$ containing $B_{R}(x_0)$. For any multi-index $\overline{n}$ define 
  \begin{align}
    r(x,x_0) &= \tilde{u}(x)-\tilde{u}(x_0)-(x-x_0)^{T} \nabla u (x_0) \\
    r_{\overline{n}}(x,x_0) &= \tilde{u}(x)-\tilde{u}(x_0)-(x-x_0)^T \nabla_{\overline{n}} u (x_0).    
  \end{align}
  
 Then, $r_{\overline{n}}(\cdot,x_0) \rightarrow r(\cdot,x_0)$ uniformly as min$(\overline{n})\rightarrow \infty$. Thus the nonlocal Taylor approximant $A_{\overline{n}}(x) = \tilde{u}(x_0)+(x-x_0)^T \nabla_{\overline{n}} u (x_0)$ also converges to the standard Taylor approximant $A(x) = \tilde{u}(x_0)+(x-x_0)^T \nabla u (x_0)$ as min$(\overline{n})\rightarrow \infty$.
\end{theorem}
 
 Thus, the nonlocal gradient obtained above allows for linear approximations similar to the usual Taylor approximant. Moreover, iterating this approach can yield higher order analogues as well. However, the ``best" linear approximation to a sufficiently smooth function will always be the one derived from the standard local-calculus based Taylor approximant.

\section{Probabilistic Reformulation, $\epsilon$-Subgradients and Biased Gradient Descent}\label{sec:bgd}
In this section, we reformulate the results of Section \ref{sec:theory} within a probabilistic framework. We also make connections with $\epsilon$-subgradients by showing that nonlocal directional derivatives are $\epsilon$-subgradients. Finally, pre-existing, well-established theoretical results of biased gradient descent apply in this case, and we therefore can obtain convergence results while using nonlocal directional derivatives in place of actual gradients while performing stochastic gradient descent-like methods.

\subsection{Probabilistic Reformulation}
Recall that the fixed sequence of densities $\{\rho_n(\cdot)\}$ are probability densities. Thus, writing out the definition of the nonlocal directional derivative, we obtain:

\begin{align}\label{eq:calcNDD}
D_{n,v}u(x) &= \int_{\mathbb{R}}\frac{u(x+tv)-u(x)}{t}\rho_n(t) dt \\
&=\int_{\mathbb{R}}\mathcal{Q}(u,v,x,t)\rho_n(t) dt \\
&=\mathbb{E}_{\rho_{n}}[\mathcal{Q}(u,v,x,T)],
\end{align}
where
\begin{equation}
\mathcal{Q}(u,v,x,t) = \frac{u(x+tv)-u(x)}{t},
\end{equation}
and the expectation $\mathbb{E}_{\rho_{n}}[u,v,x,T]$ is taken with respect to the density $\rho_n(\cdot)$ in the variable $T.$ Intuitively, this observation tells us that the nonlocal directional derivative can be numerically estimated by drawing samples $t_i, i = 1,\ldots,$ from the distribution $\rho_n(\cdot),$ evaluating the quotient $\mathcal{Q}(u,v,x,t_i)$ and performing a simple averaging. Indeed, this amounts to a Monte Carlo type evaluation of the difference quotient. The key point here is that the samples need to be drawn from the specific distribution $\rho_n(\cdot),$ and not by an indiscriminate random sampling. A prudent choice of $\rho_n(\cdot)$ is therefore critical not only for the theoretical convergence of $D_{n,v}u(x)$ to $D_{v}u(x)$, but also for numerical stability of the estimation scheme mentioned above.

Now, since we have established the fact that the random variable $\mathcal{Q}(u,v,x,T)$ has $D_{n,v}u(x)$ as its mean under the measure $\rho_n(\cdot)$, it is instructive to estimate its higher moments. We have the following result that provides such an estimate.

\begin{theorem}\label{thm:probform}
Let $u\in C_{0}^{1}(\Omega)$. The higher moments of $\mathcal{Q}(u,v,x,T)$ satisfy the following bound:
\begin{equation}\label{eq:momentbd}
|\mathbb{E}_{\rho_n}[\mathcal{Q}^k(u,v,x,T)]|\leq \|v\|^k\|\nabla u\|_{L^\infty}^{k},
\end{equation}
for $k=1,\ldots.$ In particular, the variance of $\mathcal{Q}(u,v,x,T)$ is bounded as:
\begin{equation}\label{eq:variancebd}
\text{var}_{\rho_n}[\mathcal{Q}(u,v,x,T)]\leq \|v\|^2\|\nabla u\|_{L^\infty}^{2}-(D_{n,v}u(x))^2.
\end{equation}
\end{theorem}
As expected, when $n\rightarrow \infty,$ we know $D_{n,v}u\rightarrow D_{v}u$ (when the term $D_{v}u$ exists), and hence the estimate on the variance indicates that the bound on the variance decays as $n\rightarrow \infty.$ Moreover, for the bound in Theorem \ref{thm:probform} to be meaningfully useful for large $k$, we would need either $\|\nabla u\|_{L^\infty}<1$ or $\|v\|<1,$ so that for large $k$, the right hand side of the bound stays bounded. While we have little control over $\|\nabla u\|_{L^\infty},$ we can choose $v$ such that $\|v\| = O(\frac{1}{\|\nabla u\|_{L^\infty}}),$ whence $|\mathbb{E}_{\rho_n}[\mathcal{Q}^k(u,v,x,T)]|\leq M(k),$ and $M(k)$ depends on $k$ and remains bounded as $k$ becomes large.

\subsection{Biased nonlocal gradient descent}
A natural question that arises is: what can be said of (stochastic) gradient descent-like algorithms using the nonlocal gradient in place of the actual gradient? In other words, how much of \emph{biased gradient descent} theory can be transferred to the present context? The literature on biased/imperfect stochastic gradient descent is large, and different assumptions on the biased gradient (its form, its noise etc.) yield differing estimates on the convergence of the corresponding biased gradient descent algorithms (see for e.g. \citealt{biased1,biased2} and references therein). We focus on the work of  \cite{biased2} and derive convergence estimates for stochastic gradient descent using the nonlocal gradient operators presented here. For completeness, we state the relevant results and assumptions of \cite{biased2} below.

Assume that we have a sufficient smooth function $u$ with gradient $\nabla u$. The (stochastic) biased gradient descent algorithm we consider is:

\begin{align}
x_{k+1} &= x_{k}-\gamma_{k}g_{k}(x_k) \\
g_{k}(x_k) &= \nabla u(x_k) + b(x_k)+n(x_k,\xi_k),
\end{align}
where $b(x_k)$ is the bias term and $n(x_k,\xi_k)$ is a zero mean observation noise term depending possibly on both $x_k$ and a realization of the noise $\xi$. Assume that the bias and noise terms satisfy the following conditions:

\begin{align}
\|b(x)\|^2 &\leq m\|\nabla u(x)\|^2 + \zeta^2, \, \forall x \in \mathbb{R}^N,\\
\mathbb{E}_{\xi}n(x,\xi)&=0, \, \forall x \in \mathbb{R}^N, \\
\mathbb{E}_{\xi}\|n(x,\xi)\|^2 &\leq M \|\nabla u(x)+b(x)\|^2 + \sigma^2, \, \forall x \in \mathbb{R}^N,
\end{align}

where the constants $m,M,\zeta,\sigma$ do not depend on $x$.
The authors of \cite{biased2} provide convergence estimates for $U_{k} := \mathbb{E}u(x_k) - u^{*}$, where \begin{equation}u^{*}:=\text{min}_{x\in\mathbb{R}^N}u(x),\end{equation}
where the $x_k$ are the iterates of the stochastic biased gradient descent algorithm. We use the following form of their results:

\begin{theorem}\label{thm:biasedgd}(from \cite{biased2}) Assume that the function $u$ is differentiable and satisfies the following conditions:
\begin{itemize}
\item There is a constant $L>0$ such that for all $x,y\in \mathbb{R}^N$, we have \begin{equation}u(y) \leq u(x) + \nabla u (x)^{T}\cdot (y-x) + \frac{L}{2}\|y-x\|^2\end{equation}
\item (Polyak- Lojasiewicz (PL) condition) There is a constant $\mu>0$ such that for all $x\in \mathbb{R}^N$, we have \begin{equation}\|\nabla u (x)\|^2 \geq 2\mu(u(x)-u^{*}).\end{equation}
\end{itemize}
Assume further the bounds on the bias and noise term stated earlier. Then, choosing the step size $\gamma$ appropriately ensures that for any $\epsilon \geq 0,$ we need \begin{equation}K=\mathcal{O}((M+1)\text{log}(\frac{1}{\epsilon})+\frac{\sigma^2}{\epsilon \mu (1-m)+\zeta^2})\frac{L}{\mu(1-m)}\end{equation} iterations of the stochastic biased gradient descent algorithm to ensure that \begin{equation}U_K=\mathcal{O}(\epsilon + \frac{\zeta^2}{\mu(1-m)}).\end{equation}
\end{theorem}
In the present case, we will see that the nonlocal gradient $\nabla_{\overline{n}}u$ satisfies the bounds that are required of it to serve as a biased gradient. Thus, the results of \cite{biased2} apply and we can conclude a similar result for stochastic nonlocal gradient descent. 

\begin{theorem}\label{thm:nlbiased}
Let $\epsilon>0$ be fixed, and let $\mathcal{N}(\xi)$ be random zero mean, finite variance noise independent of $x$, i.e., $\mathbb{E}[\|\mathcal{N}(\xi)\|^2]  = \sigma^2$. Then there is an $N_\epsilon>0$ such that for all $\text{min}(\overline{n})>N_{\epsilon},$ the $g(x,\xi) = \nabla_{\overline{n}} u(x)+ \mathcal{N}(\xi)$ are biased gradients.
\end{theorem}

\subsection{$\epsilon$-subgradients and $D_{n,v}u(x)$}
We now take yet another view of $D_{n,v}u(x).$ Let $\epsilon\geq 0$ be given. Recall that an $\epsilon$-subgradient of a convex function $u$ is a vector $z\in \mathbb{R}^N$ such that $u(y)-u(x) \geq z\cdot(y-x)-\epsilon, \, \forall y \in \mathbb{R}^N$ (\citealt{auslender2004interior,millan2019inexact,guo2014generalized}). The collection of all such $\epsilon$-subgradients is the $\epsilon$-subdifferential set defined as:

\begin{equation}
\partial_{\epsilon}u(x) = \{z \in \mathbb{R}^N: u(y)-u(x) \geq z\cdot(y-x)-\epsilon, \, \forall y \in \mathbb{R}^N\}.
\end{equation}
In particular, if $\epsilon = 0,$ then $\partial_{\epsilon}u(x)$ is the usual subdifferential consisting of standard subgradients. In the present context, given any $\epsilon > 0,$ we can show that there is an $n_\epsilon>0$ such that for all multi-indices $\overline{n} = (n_1,\ldots,n_N)$ with $\text{min}(\overline{n})>n_\epsilon,$ the collection $\{\nabla_{\overline{n}}u\}\subset \partial_{\epsilon}u(x).$

\begin{theorem}\label{thm:epsilonsubgrad}Let $u$ be a convex, closed function on $\Omega$ and let $\epsilon$ be arbitrary. Then, there is a $N_\epsilon>0$ such that for all multi-indices $\overline{n} = (n_1,\ldots,n_N)$ with $\text{min}(\overline{n})>N_\epsilon,$ the collection $\{\nabla_{\overline{n}}u:\text{min}(\overline{n})>n_\epsilon\}\subset \partial_{\epsilon}u(x).$
\end{theorem}
Now that we have established the fact that nonlocal gradients are $\epsilon-$subgradients, and we also know that nonlocal gradients yield biased gradients, it is but natural to ask if \emph{all} $\epsilon-$subgradients give rise to biased gradients. Indeed, this generalization is true:

\begin{theorem}\label{thm:epsilonbiased}
Let $\epsilon>0$ be fixed, $b(x)\in \partial_{\epsilon}u(x),$ and $n(x,\xi) = \mathcal{N}(\xi)$ be random zero mean, finite variance noise independent of $x$, i.e., $\mathbb{E}[\|\mathcal{N}(\xi)\|^2]  = \sigma^2$. Then, $g(x,\xi) = \nabla u(x) + b(x) + \mathcal{N}(\xi)$ is a biased gradient.
\end{theorem}
Thus, any $\epsilon-$subgradient is a biased gradient and as such can be used in first order stochastic optimization algorithms.

\subsection{Convergence with random directions}
In this section, we consider the randomization of the direction $v$ used in calculating $D_{n,v}u(x)$. Given a distribution $\mathcal{P}_V$ on $\mathbb{R}^N,$ consider a random direction $v\sim \mathcal{P}_V$ drawn from the distribution $\mathcal{P}_V$. From what we have thus seen, for a suitable function $u$, we know that $D_{n,v}u(x)\rightarrow D_{v}u(x) = v^{T}\nabla u(x)$ in some topology. We thus expect that \begin{equation}\mathbb{E}_{v\sim \mathcal{P}_V}[D_{n,v}u(x)]\rightarrow \mathbb{E}_{v\sim \mathcal{P}_V}[v]^{T}\nabla u(x).\end{equation}
The objective of this section is to prove this result under some mild conditions on the distribution $\mathcal{P}_V,$ and specific choices of $\rho_n(\cdot).$ 

\begin{theorem}\label{thm:randomdirconv}
Assume that $u\in C^{1}_{0}(\Omega$) and that $\mathcal{P}_V$ has norm-bounded moments, i.e., $\int_{\mathbb{R}^n}\|v\|^k d\mathcal{P}_V<\infty$ for all finite $k$. Choose $\rho_n(t) = n(n-1)\mathbb{I}_{[\frac{1}{n},\frac{1}{n-1}]}.$ Then \begin{equation}\|\mathbb{E}_{v\sim\mathcal{P}_{V}}[D_{n,v}u - D_{v}u]\|_{L^{\infty}}\rightarrow 0\end{equation} as $n\rightarrow \infty.$
\end{theorem}
The choice of $\rho_n(\cdot)$ above is not unique. Another possible choice to ensure the same result is $2^n\mathbb{I}_{[\frac{1}{2^n},\frac{1}{2^{n-1}}]}.$ The choice in Theorem \ref{thm:randomdirconv} allows for easier bounds.
In the following section, we use the tools developed thus far to bear on the interesting case of Brownian motion.

\section{Brownian Activation}\label{sec:brownian}
In this section, we consider the sample paths of Brownian motion. Recall (\citealt{shreve}) that Brownian motion is a Gaussian process characterized by independent increments, and whose sample paths are $\alpha$-H\"older continuous for all $\alpha<\frac{1}{2}$ and not $\alpha$-H\"older continuous for all $\alpha>\frac{1}{2}.$ We fix the interaction kernel sequence $\rho_n(\cdot)$ in this section to be the exponential moving rectangles $\rho_{n}(t) = 2^{n}\mathbb{I}_{[\frac{1}{2^{n}},\frac{1}{2^{n-1}}]}(t).$

Let $W(x,\omega) = W(x)$ denote standard one-dimensional Brownian motion, where $\omega$ refers to a particular random sample leading to a sample path. From the results presented earlier, we know that the sample paths of $W(x)$ have well defined NDD $D_{n,v} W(x).$ In the one-dimensional case, $v\in \mathbb{R}.$ We then define the process $D_{n,v} W(x,\omega)$ sample path-wise. It is clear that $D_{n,v} W(x)$ is Gaussian. We can also compute its mean and variance:

\begin{theorem}\label{thm:brownian}
$D_{n,v} W(x)$ is a zero-mean Gaussian process with variance given by \begin{equation}2^{n+1}|v|(1-\text{ln}(2)).\end{equation}
\end{theorem}
As expected, the variance of $D_{n,v} W(x)\rightarrow \infty$. Indeed, otherwise, this would imply that $D_{n,v}W(x)$ converges (for some $x$) to a finite value, contradicting the strong non-differentiability of sample paths. With a well-defined NDD of Brownian sample paths, we can proceed to using Brownian sample paths in activation functions of select neurons in any deep learning architecture. Consider a deep NN with ``nondiff" neurons having ``nondiff" activation functions of the form $ReLU(x) + \alpha W(x)$. Continuing as ``usual", we can train such a model, except that we would use the NDD during training. Specifically, we first fix $n$ (although in practice, this could be a tunable network parameter) and select a subset of neurons to have Brownian sample path-based activations. The use of these nondiff neurons proceeds in the standard way, with the exception that during backpropagation of gradients, the nondiff activation function and its NDD are used in place of a standard activation and the corresponding standard gradient for these neurons. No changes are made to the other ``usual" neurons. The location of nondiff neurons in the network is also a tunable parameter. Concretely, if $\phi(x) = ReLU(x) + \alpha W(x)$ is a nondiff neuron's activation function with a tunable $\alpha>0$, then its forward propagation of a feature $\textbf{z}$ is defined componentwise by $ReLU(\textbf{z}) + \alpha W(\textbf{z})$ while its backpropagated gradient is $H(\textbf{z}) + \alpha D_{n,v} W(\textbf{z}),$ where $H(\textbf{z})$ is a given implementation of the gradient of $ReLU(\cdot).$ Since we are interested in only the sample values of the activation and its NDD at $\textbf{z}$, we use the unnormalized density (i.e. without simulating the entire path) of $W(\cdot)$ and $D_{n,v} W(\cdot):$ \begin{align}W(x)&\sim \mathcal{N}(0,x)\\D_{n,v} W(x)&\sim \mathcal{N}(0,2^{n+1}\beta),\end{align} 
where $\beta = |v|(1-\text{ln}(2)).$ We now consider numerical experiments using the theory developed thus far.

\section{Experiments}\label{sec:experiments}
\begin{figure}[h]
    \centering
    \includegraphics[width=0.75\linewidth]{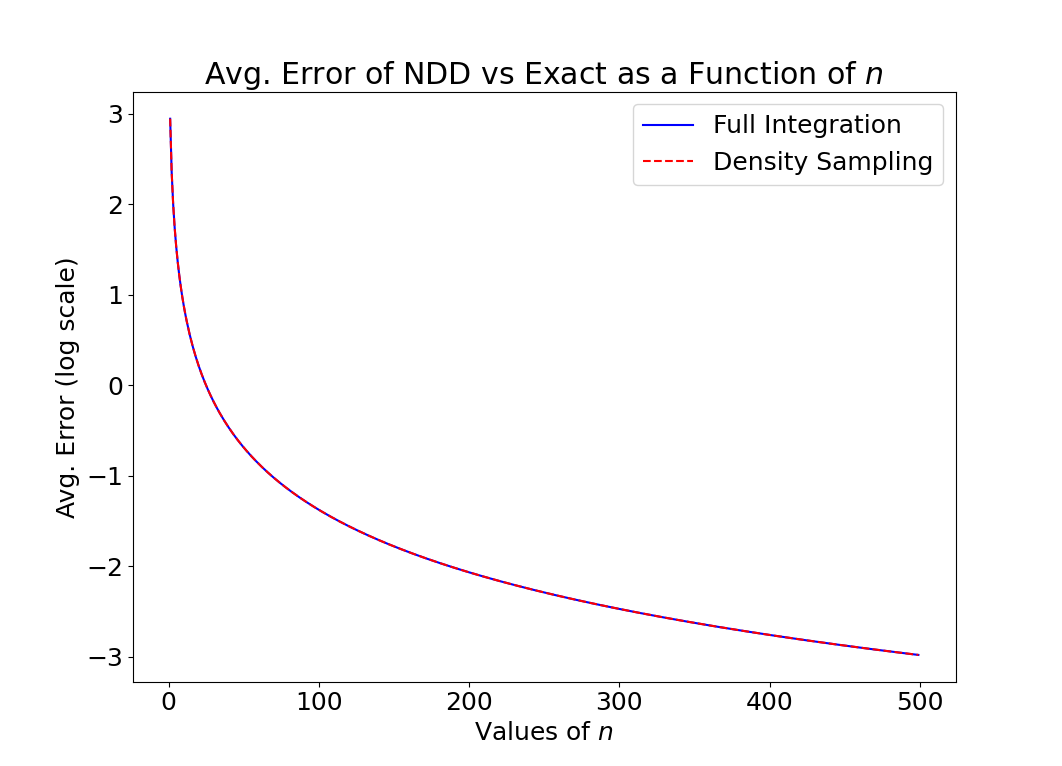}
    \caption{Convergence of NDD to exact derivative}
    \label{fig:figure1}
\end{figure}

We first begin with some ``sanity checks" with regard to calculating the NDD. Recall that we have two different ways of computing a NDD: i) evaluating the integral in Equation \ref{eq:calcNDD} via exact quadrature based integration, or sampling from the density $\rho_n(\cdot)$ and evaluating the expectation in \ref{eq:calcNDD} by averaging. In the first experiment, we consider a quadratic function of the form $\frac{1}{2} x^{T} A x + x^{T}b,$ where $A$ is symmetric. We fix the dimension of the space to be $50$, generate $250$ random $x$-points, a random unit-vector $v$, and compare the error (averaged across the different points) between the NDD at these points against the exact directional derivative $v^T Ax+v^T b.$ We compare both methods of evaluating the NDD in Figure \ref{fig:figure1}. As a function of $n=1,\ldots,500$, we see that the error rapidly decays (the errors are plotted on a logarithmic scale), and there is no appreciable difference in the computation of the NDD between the complete integral vs sampling approach. Thus we see that a nominal choice of $n\sim 10$ would suffice in practice. 
\subsection{Translating Disc IAM}
\begin{figure}[h]
    \centering
    \includegraphics[width=0.75\linewidth]{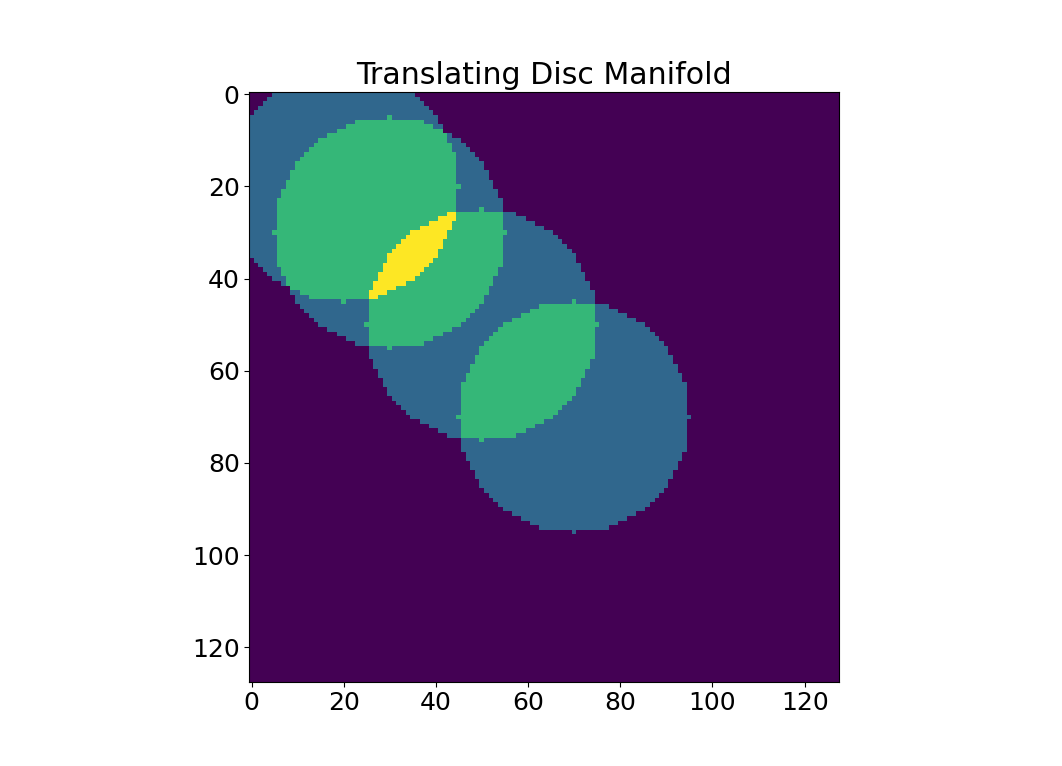}
    \caption{Translating Disc Manifold}
    \label{fig:figure2}
\end{figure}

\begin{figure}[h]
    \centering
    \includegraphics[width=0.75\linewidth]{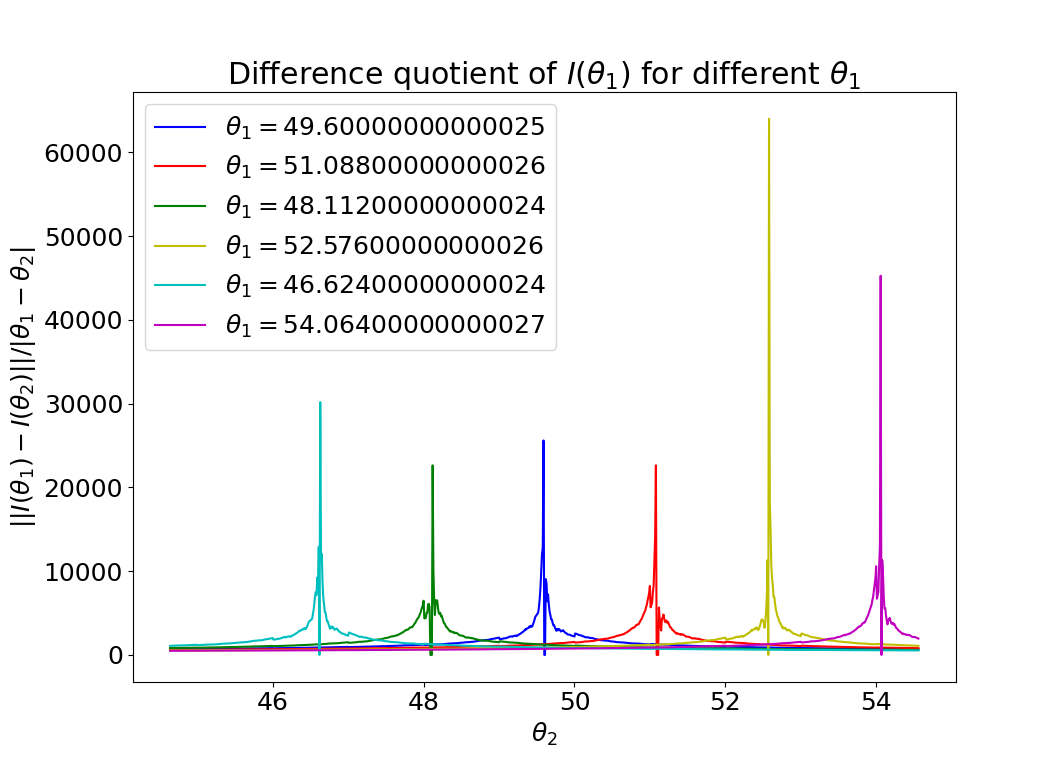}
    \caption{Nowhere Differentiability of $I_\theta$}
    \label{fig:figure3}
\end{figure}

\begin{figure}[h]
  \centering
  \begin{minipage}{0.48\textwidth}
    \centering
    \includegraphics[width=\linewidth]{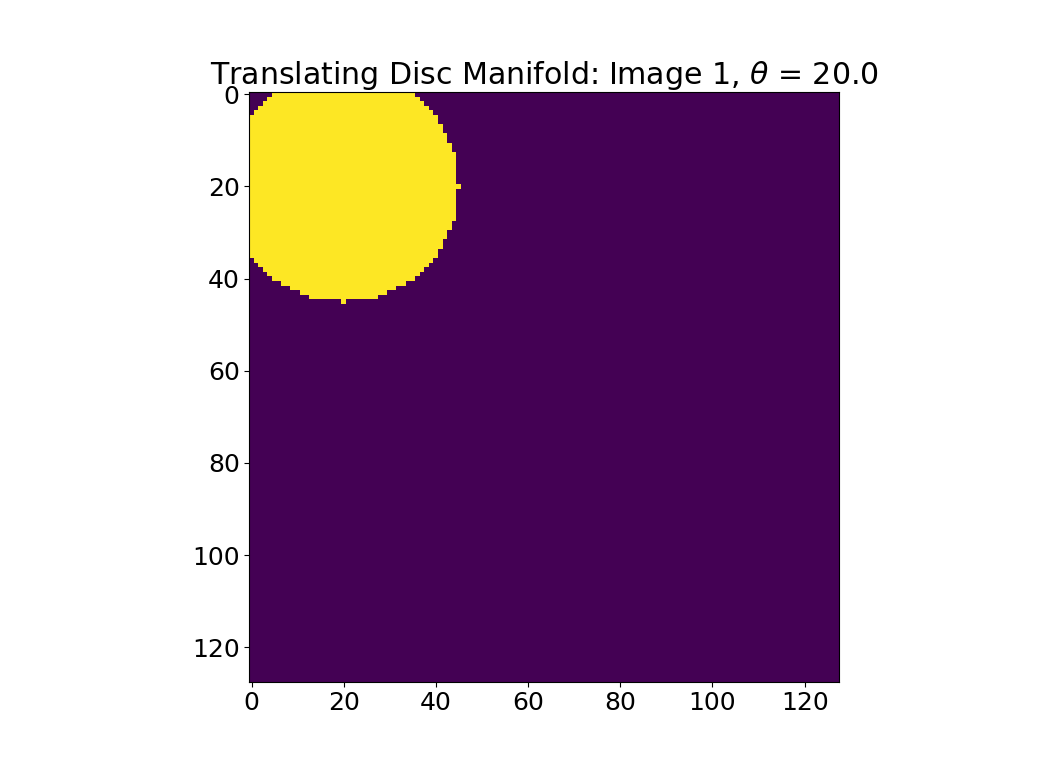}
    \caption{Image $I_{\theta}$}
    \label{fig:figure4a}
  \end{minipage}
  \hfill
  \begin{minipage}{0.48\textwidth}
    \centering
    \includegraphics[width=\linewidth]{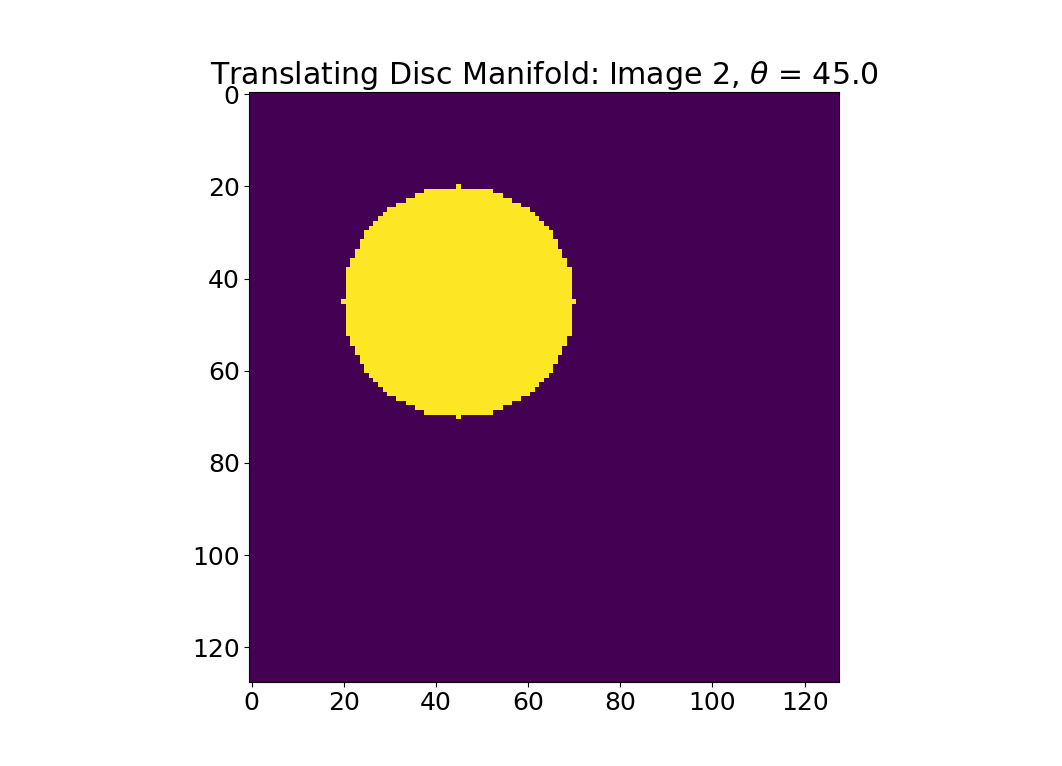}
    \caption{Image $I_{\theta^{*}}$}
    \label{fig:figure4b}
  \end{minipage}
  \caption{Estimating $\theta^{*}$ given $I_{\theta}$ and $I_{\theta^{*}}$}
  \label{fig:figure4}
\end{figure}

We consider now a highly nondifferentiable and nonconvex problem of parameter estimation on a nonsmooth image manifold. This example is motivated by the results in \cite{wakin2005high} and \cite{wakin2005multiscale}. Briefly, given a collection of images obtained by imaging a scene from a variety of vantage points, the collection of images lies on a low-dimensional manifold (called an image articulation manifold or IAM) in the large dimensional space of all images of a given resolution. As an example, we look at the manifold $\mathcal{M}$ of images of a translating disc (Figure \ref{fig:figure2}). This is a 1-dimensional manifold corresponding to the translation of the disc from the top left corner to the bottom right corner. Each image $I=I_\theta$ on the manifold (i.e. curve) corresponds to a translation $\theta\in \Theta = [0,1],$ where $\theta=0$ corresponds to the image $I_0$ of the disc at the top left corner. Thus we have a bijective map $\phi:\Theta \rightarrow \mathcal{M}$ given by $\theta\mapsto I_\theta.$ Hence, $\mathcal{M}=\phi(\Theta)=\{I_{\theta}:\theta\in\Theta\}.$ It is well-known (see \citealt{wakin2005high} and \citealt{wakin2005multiscale}) that the map $\theta\mapsto I_\theta$ is severely nondifferentiable, indeed, $\frac{\|I_{\theta_{1}}-I_{\theta_{2}}\|}{|\theta_1-\theta_2|}\sim C|\theta_1-\theta_2|^{-\frac{1}{2}}$. As shown in Figure \ref{fig:figure3}, the map is nowhere differentiable.  However, being H\"older continuous (with $\alpha=\frac{1}{2}$), the map's NDD does exist.
The problem we wish to solve, then, is given an image $g=I_{\theta^{*}},$ recover an estimate $\hat{\theta}$ of the the parameter (i.e. translation) $\theta^{*}$. The situation is illustrated in Figure \ref{fig:figure4}. We formulate this (following \citealt{fods} and \citealt{wakin2005multiscale}) as the problem of finding $\hat{\theta} = \text{argmin}_{\theta}\mathcal{E}(\theta) $ where $\mathcal{E}(\theta):= \|I_\theta-g\|.$ Unlike the approach taken in \cite{wakin2005multiscale} which relies on smoothing the manifold $\mathcal{M},$ we follow the the approach taken in \cite{fods} and minimize $\mathcal{E}(\theta)$ directly using gradient descent. Thus, we perform $\theta_{k+1} = \theta_{k}-\gamma_{k} \nabla_{n} \mathcal{E}(\theta_k),$ where $\nabla_{n} \mathcal{E}(\theta_k)$ is evaluated via sampling as discussed earlier. In practice, given the severe nonconvexity of the objective function, we resort to re-initializing the guess when the NDD does not yield a descent direction.

\begin{figure}[h]
  \centering
  \begin{minipage}{0.48\textwidth}
    \centering
    \includegraphics[width=\linewidth]{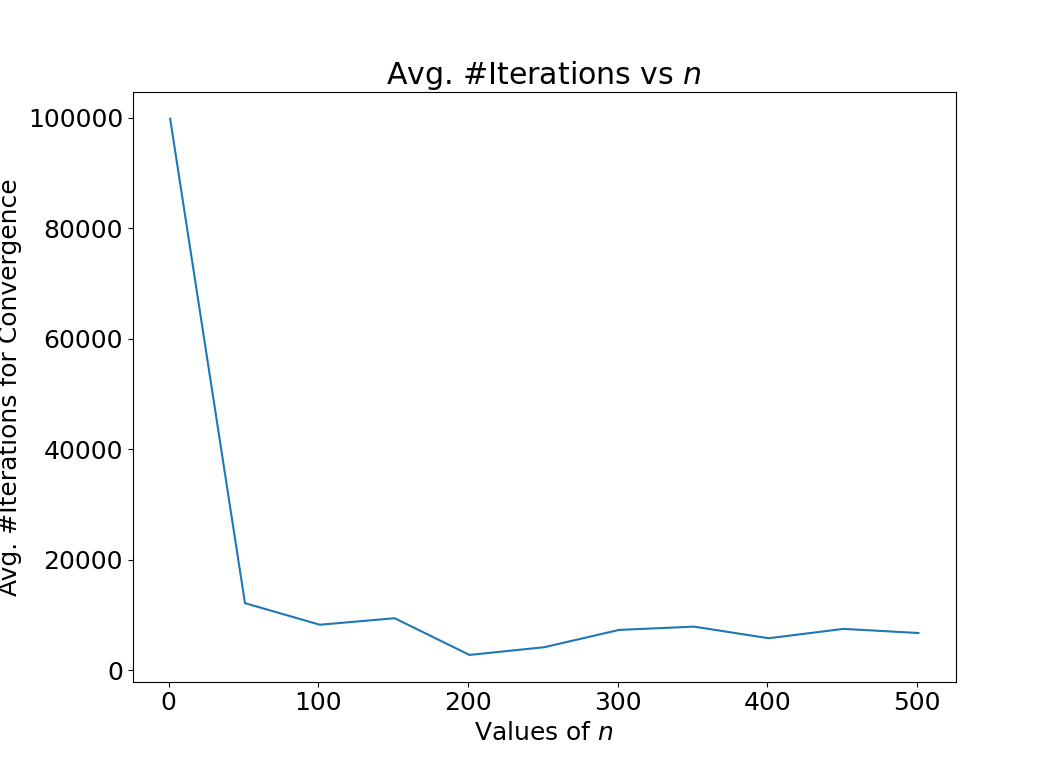}
    \caption{Avg. iterations vs $n$}
    \label{fig:figure5a}
  \end{minipage}
  \hfill
  \begin{minipage}{0.48\textwidth}
    \centering
    \includegraphics[width=\linewidth]{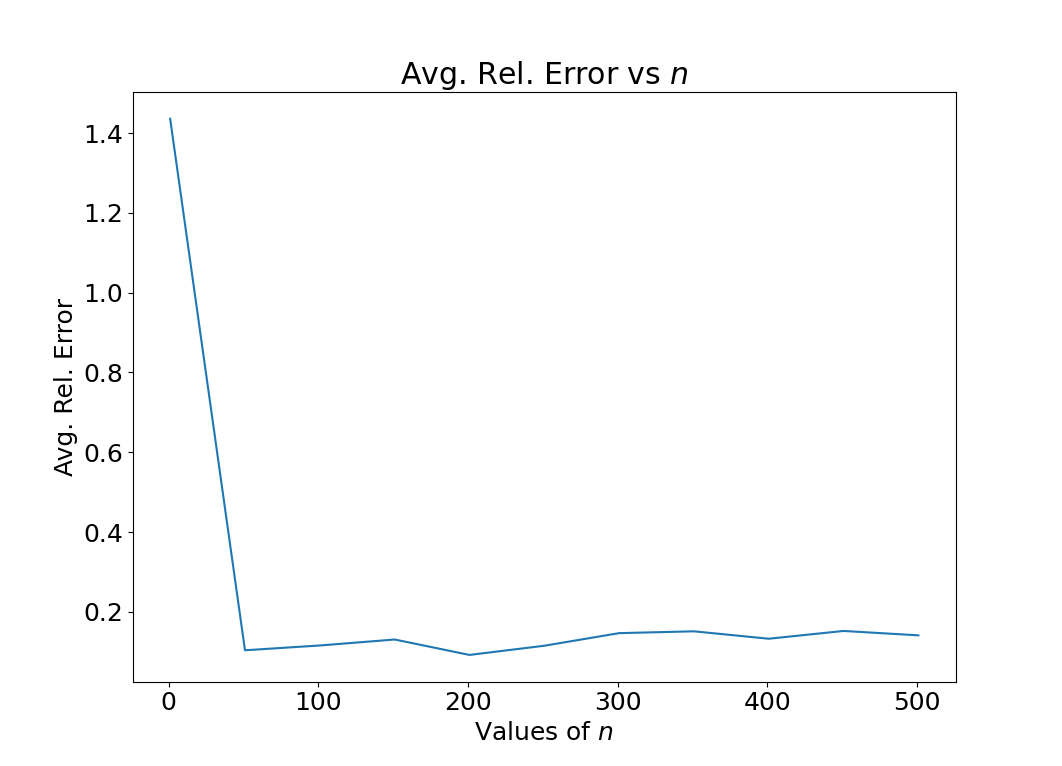}
    \caption{Avg. error vs $n$}
    \label{fig:figure5b}
  \end{minipage}
  \caption{Avg. iterations and error as a function of $n$}
  \label{fig:figure5}
\end{figure}

We solve the problem for a range of $n$ with $1000$ different runs having randomized starting images and the moving rectangle family for $\rho_n(\cdot).$ We average out the results over the different runs and plot the average number of iterations and average relative error between the estimated $\hat{\theta}$ and true $\theta^{*}$. The results are seen in Figure \ref{fig:figure5}. We see that we have an average relative error $\sim 0.1$ once $n>100$ while the average number of iterations for convergence stabilizes to $\sim 15000$. We are able to thus efficiently solve this highly nondifferentiable and nonconvex problem using the proposed approach. 

\subsection{Image Classification}
We now consider the example of image classification using standard deep learning architectures. The underlying hypothesis is that the use of nondiff neurons will naturally prevent overfitting much akin to dropout. Thus, the questions to be investigated are i) when training with lower amount of data, does the use of nondiff neurons help the accuracy/prevent overfitting? ii) is there any pattern in the location of nondiff neurons within the network? iii) what role does the variance of Brownian motion play? To answer these questions, we tested our approach first on the extended MNIST (E-MNIST) dataset. We used a standard small multilayer perceptron (MLP) followed by an adversarial MLP which is deeper and therefore prone to being overfit. All neurons in a specific layer have ReLU replaced with Brownian ReLU. We varied the location of nondiff neurons across layers as well as the scaling ($\alpha$, see Section \ref{sec:brownian} for details about this parameter) parameter but kept the variance fixed. Finally, we varied the percentage of the dataset used for training, with a fixed split for each seed.

\subsubsection{Standard MLP}

\begin{figure}[h]
  \centering
  \begin{minipage}{0.48\textwidth}
    \centering
    \includegraphics[width=\linewidth]{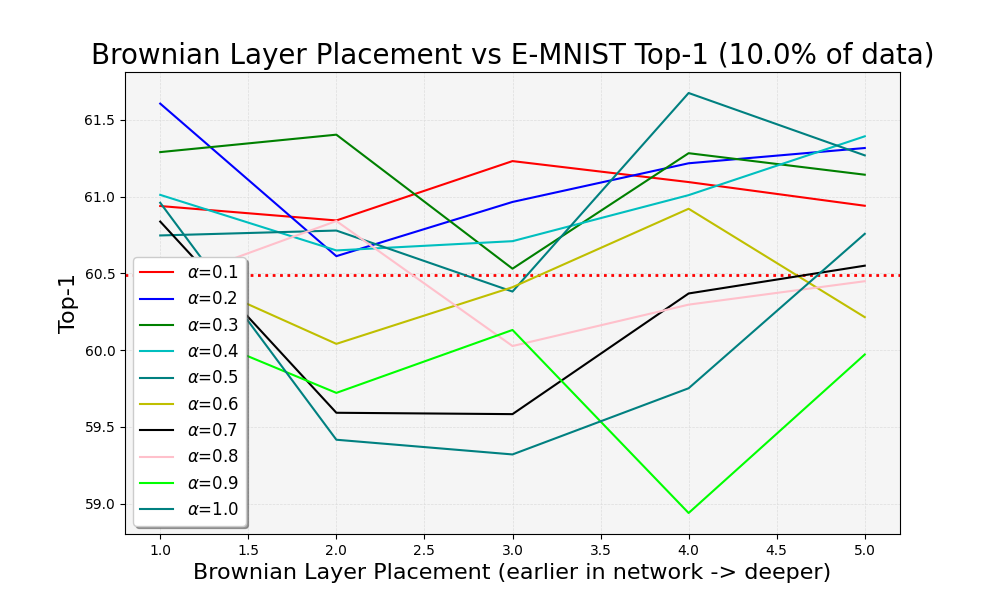}
    \caption{Low Training Data Regime}
    \label{fig:figure6a}
  \end{minipage}
  \hfill
  \begin{minipage}{0.48\textwidth}
    \centering
    \includegraphics[width=\linewidth]{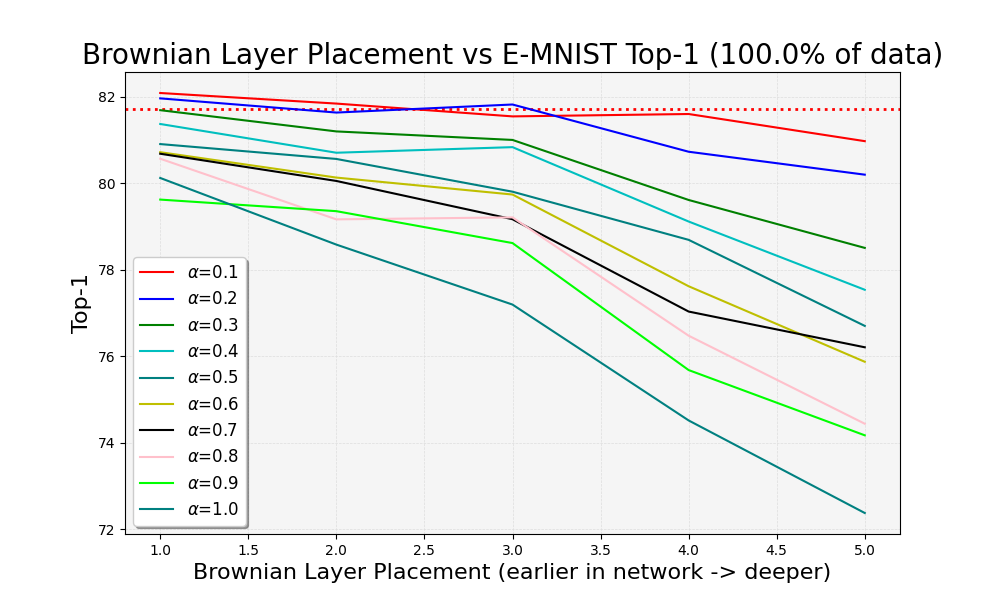}
    \caption{Second figure}
    \label{fig:figure6b}
  \end{minipage}
  \caption{High Training Data Regime}
  \label{fig:figure6}
\end{figure}

The baseline MLP architecture ([neurons x layers]) was [128x3, 64x3]. We found that in the low training data regime, better relative Brownian performance was observed (\ref{tab:emnist_basic}). The architecture consistently favors lower $\alpha$ across data both low and high training data regimes. In terms of the location of nondiff neurons, in low data regimes, there was no systematic relationship between location and performance while when using all of the data for training, there was a strong preference for nondiff neurons to be positioned earlier in the network for higher accuracy. This intuitively seems consistent: using nondiff neurons in earlier parts of the network promotes ``exploration" of the loss landscape, and using standard ReLU neurons in the later stages promotes ``exploitation". This is seen in Figure \ref{fig:figure6}.

\subsubsection{Adversarial MLP}
In the ``adversarial" network, which was susceptible to overfitting, the observed gap between nondiff and regular (baseline) neuron models widents dramatically (\ref{tab:emnist_adv}). The tendency towards lower values of $\alpha$ disappears, while the nondiff placement trend remains inconclusive. 

\begin{table}[h]
\centering
\begin{tabular}{cccc}
\textbf{Model} & \textbf{Data Pct (\%)} & \textbf{Top-1 (\%)} & \textbf{Top-3 (\%)} \\
Baseline & 10 & 60.5 & 80.6 \\
Brownian & 10 & 62.1 & 81.2 \\
Baseline & 50 & 62.2 & 79.8 \\
Brownian & 50 & 63.0 & 80.6 \\
Baseline & 100 & 81.7 & 95.2 \\
Brownian & 100 & 82.1 & 95.4 \\
\end{tabular}
\caption{E-MNIST accuracies of baseline (no Brownian layers) \textbf{standard} MLP performance versus top-performing \textbf{standard} MLP with a Brownian layer. "Data Pct" refers to the percentage of the dataset used for training. Results are averaged across 3 seeds.}
\label{tab:emnist_basic}
\end{table}

\begin{table}[h]
\centering
\begin{tabular}{cccc}
\textbf{Model} & \textbf{Data Pct (\%)} & \textbf{Top-1 (\%)} & \textbf{Top-3 (\%)} \\
Baseline & 10 & 24.0 & 50.6 \\
Brownian & 10 & 39.4 & 64.0 \\
Baseline & 50 & 36.7 & 59.3 \\
Brownian & 50 & 51.3 & 72.0 \\
Baseline & 100 & 64.3 & 86.4 \\
Brownian & 100 & 70.7 & 89.8 \\
\end{tabular}
\caption{E-MNIST accuracies of baseline (no Brownian layers) \textbf{adversarial} MLP versus top-performing \textbf{adversarial} MLP with a Brownian layer. "Data Pct" refers to the percentage of the dataset used for training. Results are averaged across 3 seeds.}
\label{tab:emnist_adv}
\end{table}

\subsection{Text Classification Fine-Tuning}
As a final example illustrating the use of nondiff neurons, we consider text classification. Since we are interested in low data regimes, we consider the task of LoRA (Low-Rank Adaptation of Large Language Models) fine-tuning on a small dataset (\ref{tab:roberta}). While the use of nondiff neurons helps to increase accuracy on some benchmarks, there are little improvements over the baseline on others. Surprisingly, we observed no correlation between brownian performance and size of fine-tuning dataset. As before, we varied $\alpha$ and the nondiff layer placement. We observed that the network favored nondiff neurons placed in later layers (layer 7/12 or layer 10/12 performed best). Again, similar to the basic MLP case, we saw that low $\alpha$ was preferred, with the best results coming from an $\alpha = 0.01$.
\begin{table}[h]
\centering
\begin{tabular}{ccccccc}
\textbf{Model} & \textbf{MRPC} & \textbf{RTE} & \textbf{CoLA} & \textbf{QNLI} & \textbf{STSB} & Avg \\
Baseline & 90.1 & 86.5 & 63 & 92.8 & 91.7 & 84.8 \\
Brownian & 90.5 & 86.4 & 64.3 & 92.9 & 91.9 & 85.2 \\
\end{tabular}
\caption{GLUE benchmark performances of base (no Brownian layers) model versus top-performing model with a Brownian layer, fine-tuned with LoRA. Note that each GLUE task may use a different metric. Results are averaged across 3 seeds.}
\label{tab:roberta}
\end{table}

\section{Conclusions}
In this paper, we systematically tackled the concept of nonlocal directional derivatives (NDDs), and showed that it is applicable to a large class of nondifferentiable functions. In particular, ordinarily nowhere differentiable H\"older continuous functions have finite NDDs. We also showed how NDDs can be computed with sampling methods, and that NDDs are in fact $\epsilon-$ subgradients. As an extreme use case of our methodology, we successfully solved the image articulation manifold (IAM) problem of parameter estimation with a highly nondifferentiable and nonconvex objective function. We then applied our theory to study the use of NDDs in training deep learning models driven by nondifferentiable (``nondiff") activation functions, in particular, those derived from sample paths of Brownian motion. We found that in low training data regimes, the use of nondiff neurons helps in generalization and preventing overfitting. Moreover, the trends we found were seen across disparate data modalities (image and language).

While we surmounted the issue of direct evaluation of NDDs using sampling methods from a fixed class of densities, a future direction of work could address the issue of optimal classess of densities for a given problem. Indeed, depending on the optimization problem, it is conceivable that there would be an optimal class of interaction kernel densities that deliver the fastest convergence. Another possibility for future work would be extending the results presented herein to the cases of i) infinite dimensional optimization problems and ii) nonlinear domains i.e. curved manifold domains. Finally, there are open questions regarding the use of biased gradients for physics informed deep learning models. NDDs could be one possible way forward in that regard.



\acks{Work done at Southwest Research Institute (SwRI), San Antonio, Texas, USA.}


\newpage

\appendix
\section{Proofs}\label{sec:proofs}
In this section, the inner product between vectors $v,w\in \mathbb{R}^N$ will be denoted by $v\cdot w.$ The vector $(l^p)$ norm will be denoted as $\|x\|_p= \sum_{i=1}^{N}|x_i|^p,$ for $1\leq p < \infty$ and $\|x\|_\infty = \text{max}\{|x_i|\}.$ When written without a subscript, $\|x\|$ shall denote the standard Euclidean $l=2$ vector norm. The symbol $\mathcal{S}(X)$ will denote the standard Schwartz class of rapidly decaying functions on $X.$
\paragraph{Proof of Proposition \ref{thm:3density}}
\paragraph{Gaussian:} The Gaussian $\rho_n(\cdot)$ family can be realized as a sclaed version of the standard normal probability density. Hence, this family represents a nascent Dirac delta and converges in the sense of distributions to the Dirac delta.
\paragraph{Exponential Moving Rectangles:} This is an asymmetric family $\rho_n(-t) \neq \rho_n(t).$ We prove that for any $\phi\in C_{0}^{\infty}(\mathbb{R}),$ we have that \begin{equation}\langle \phi, \rho_n \rangle := \int \phi(t)\rho_{n}(t) dt \rightarrow \phi(0) = \langle \phi, \delta_0 \rangle. \end{equation}
Now, \begin{align}\int \phi(t)\rho_{n}(t) dt &= 2^{n} \int_{\frac{1}{2^n}}^{\frac{1}{2^{n-1}}}\phi(t) dt \\&= 2^{n}(\frac{1}{2^{n-1}}-\frac{1}{2^n})\phi(\xi_n) = \phi(\xi_n),\end{align}
where the second equality is by the mean value theorem and $\xi_n\in [\frac{1}{2^{n}},\frac{1}{2^{n-1}}]$ is some intermediate value. Since $\frac{1}{2^{n}}\leq \xi_n \leq \frac{1}{2^{n-1}},$ we have that $\xi_n\rightarrow 0$ as $n\rightarrow \infty,$ and by continuity of $\phi,$ we have $\phi(\xi_n)\rightarrow \phi(0).$ Putting all this together, we have that $\langle \phi, \rho_n \rangle \rightarrow \langle \phi, \delta_0 \rangle.$
\paragraph{Linear Moving Rectangles:} The same argument as the exponential moving rectangle applies here, with the only change being the use of the intermediate value theorem as 
\begin{equation}\int \phi(t)\rho_{n}(t) dt = n(n-1) (\frac{1}{{n-1}}-\frac{1}{n})\phi(\xi_n) = \phi(\xi_n),\end{equation}
with $\xi_n\in [\frac{1}{{n}},\frac{1}{{n-1}}]\rightarrow 0, $ as $n\rightarrow \infty$.

\paragraph{Proof of Lemma \ref{thm:pvi}}
\begin{proof}

\textbf{Case 1: $u\in C_{0}^{1}(\Omega)$}

Let $u\in C_{0}^{1}(\Omega)$. Then, pointwise, for any $x,v\in \mathbb{R}^N$ and $t\in \mathbb{R}$, we have

\begin{align*}
u(x+tv) - u(x) &= \int_{0}^{1} tv\cdot \nabla u (x+tsv) ds \\
&=t\int_{0}^{1} v\cdot \nabla u (x+tsv) ds \\
\Rightarrow |\frac{u(x+tv) - u(x)}{t}|&= |\int_{0}^{1} v\cdot \nabla u (x+tsv) ds|\\
&\leq \int_{0}^{1} \|v\|\|\nabla u (x+tsv)\| ds\\
&\leq \|v\|\|\nabla u \|_{L^{\infty}}.
\end{align*}
Thus
\begin{align*}
|D_{\rho,v}u(x)| &\leq \text{lim}_{\epsilon\rightarrow 0} |\int_{\mathbb{R}-B_{\epsilon}(0)}\frac{u(x+tv)-u(x)}{t}\rho(t)dt| \\
&\leq \|v\|\|\nabla u \|_{L^{\infty}} \text{lim}_{\epsilon\rightarrow 0}\int_{\mathbb{R}-B_{\epsilon}(0)} \rho(t)dt\\
&\leq \|v\|\|\nabla u \|_{L^{\infty}},
\end{align*}
since $\int_{\mathbb{R}-B_{\epsilon}(0)} \rho(t)dt \leq 1.$
Thus, $|D_{\rho,v}u(x)|\leq \|v\|\|\nabla u \|_{L^{\infty}}.$

\textbf{Case 2: $u\in W^{1,p}(\Omega)$}

Let $u\in W^{1,p}(\Omega)$. Then, we can write for any $v\in \mathbb{R}^N$ and $t\in \mathbb{R}$,

\begin{align*}
\int_{\mathbb{R}^N}|u(x+tv) - u(x)|^{p}dx &\leq \|tv\|^{p}\|u\|_{W^{1,p}}^{p} \\
\Rightarrow  \int_{\mathbb{R}^N}|\frac{u(x+tv) - u(x)}{t}|^{p}dx &\leq \|v\|^{p}\|u\|_{W^{1,p}}^{p}.
\end{align*}
Now, we have

\begin{align*}
\|D_{\rho,v}u\|_{L^p}^{p} & =  \int_{\mathbb{R}^N}|\text{lim}_{\epsilon\rightarrow 0} \int_{\mathbb{R}-B_{\epsilon}(0)}\frac{u(x+tv) - u(x)}{t}\rho(t)dt|^{p} dx \\
&\leq_{\text{triangle ineq}} \int_{\mathbb{R}^N}\text{lim}_{\epsilon\rightarrow 0} \int_{\mathbb{R}-B_{\epsilon}(0)}|\frac{u(x+tv) - u(x)}{t}\rho(t)dt|^{p} dx \\
&\leq_{\text{H\"older ineq}} \int_{\mathbb{R}^N}\text{lim}_{\epsilon\rightarrow 0} \int_{\mathbb{R}-B_{\epsilon}(0)}|\frac{u(x+tv) - u(x)}{t}dt|^{p}\rho(t)dt  dx \\
&\leq_{\text{Fubini}} \text{lim}_{\epsilon\rightarrow 0} \int_{\mathbb{R}-B_{\epsilon}(0)}\int_{\mathbb{R}^{N}}|\frac{u(x+tv) - u(x)}{t}dt|^{p}\rho(t)dxdt \\
&\leq \text{lim}_{\epsilon\rightarrow 0} \int_{\mathbb{R}-B_{\epsilon}(0)} \|v\|^{p}\|u\|_{W^{1,p}}^{p} \rho(t) dt\\
&= \|v\|^{p}\|u\|_{W^{1,p}}^{p}.
\end{align*}
Thus 
$\|D_{\rho,v}u\|_{L^p}\leq \|v\|\|u\|_{W^{1,p}}$

\textbf{Case 3: $u\in \text{Lip}(\Omega,M)$}

Let $u\in \text{Lip}(\Omega,M)$. Then, we can write for any $v\in \mathbb{R}^N$ and $t\in \mathbb{R}$,
\begin{align*}
|u(x+tv)-u(x)|&\leq M\|tv\| \\
\Rightarrow  \frac{|u(x+tv)-u(x)|}{|t|} &\leq M\|v\|.
\end{align*}
Thus,
\begin{align*}
|D_{\rho,v}u(x)| &\leq \text{lim}_{\epsilon\rightarrow 0} |\int_{\mathbb{R}-B_{\epsilon}(0)}\frac{u(x+tv)-u(x)}{t}\rho(t)dt| \\
&\leq \text{lim}_{\epsilon\rightarrow 0} \int_{\mathbb{R}-B_{\epsilon}(0)}|\frac{u(x+tv)-u(x)}{t}\rho(t)|dt
&\leq M\|v\|\text{lim}_{\epsilon\rightarrow 0}\int_{\mathbb{R}-B_{\epsilon}(0)} \rho(t)dt. \\
&\leq M\|v\|.
\end{align*}

\textbf{Case 4: $u\in \text{H\"ol}(\Omega,M,\alpha), \, \alpha>0$ and $\frac{\rho(t)}{|t|^{1-\alpha}}$ is integrable on $\mathbb{R}$}
\end{proof}

Let $u\in \text{H\"ol}(\Omega,M,\alpha), \, \alpha>0.$ 

Then, we can write for any $v\in \mathbb{R}^N$ and $t\in \mathbb{R}$,
\begin{align*}
|u(x+tv)-u(x)|&\leq M\|tv\|^\alpha \\
\Rightarrow  \frac{|u(x+tv)-u(x)|}{|t|^\alpha} &\leq M\|v\|^\alpha.
\end{align*}

Thus,
\begin{align*}
|D_{\rho,v}u(x)| &\leq \text{lim}_{\epsilon\rightarrow 0} |\int_{\mathbb{R}-B_{\epsilon}(0)}\frac{u(x+tv)-u(x)}{t}\rho(t)dt| \\
&\leq \text{lim}_{\epsilon\rightarrow 0} \int_{\mathbb{R}-B_{\epsilon}(0)}|\frac{u(x+tv)-u(x)}{t}\rho(t)|dt \\
&\leq M\|v\|^\alpha\text{lim}_{\epsilon\rightarrow 0}\int_{\mathbb{R}-B_{\epsilon}(0)} \frac{\rho(t)}{|t|^{1-\alpha}}dt. \\
&\leq ML\|v\|^{\alpha}, \, L = \int_{\mathbb{R}} \frac{\rho(t)}{|t|^{1-\alpha}}dt < \infty.
\end{align*}
Note that if $\text{supp}(\rho(\cdot)) \subset [-A,A] \subset \mathbb{R},$ then

\begin{align*}
L &= \int_{-A}^{A} \frac{\rho(t)}{|t|^{1-\alpha}}dt\\
&\leq 2\|\rho\|_{L^{\infty}} \int_{0}^{A} \frac{1}{|t|^{1-\alpha}}dt\\
&= 2\frac{\|\rho\|_{L^{\infty}}A^{\alpha}}{\alpha},
\end{align*}
so that $|D_{\rho,v}u(x)| \leq 2\frac{M\|v\|^{\alpha}\|\rho\|_{L^{\infty}}A^{\alpha}}{\alpha}$


\paragraph{Proof of Lemma \ref{thm:linearity}}
 \begin{proof}
Linearity in $u$ is evident from the definition of $D_{\rho,v}u(x)$.

Now,

\begin{align*}
D_{\rho,v}u_1(x)u_2(x) &= \int_{\mathbb{R}} \frac{u_1(x+tv)u_2(x+tv)-u_1(x)u_2(x)}{t}\rho(t)dt \\
&=\int_{\mathbb{R}} \frac{u_1(x+tv)u_2(x+tv)-u_1(x)u_2(x+tv)+ u_1(x)u_2(x+tv) - u_1(x)u_2(x)}{t}\rho(t)dt\\
&=\int_{\mathbb{R}}\frac{u_1(x+tv)-u_1(x)}{t}u_2(x+tv)\rho(t)dt + u_1(x)D_{\rho,v}u_2(x).
\end{align*}
By symmetry, \begin{equation*}D_{\rho,v}u_1(x)u_2(x)  = \int_{\mathbb{R}}\frac{u_2(x+tv)-u_2(x)}{t}u_1(x+tv)\rho(t)dt + u_2(x)D_{\rho,v}u_1(x).\end{equation*}

Thus, adding the equalities stated above, we get

\begin{equation*}D_{\rho,v}u_1(x)u_2(x) = \frac{1}{2}(u_1 D_{v}(u_2)(x)+u_2 D_{v}(u_1)(x) + b_{\rho,v}(u_1,u_2),\end{equation*}
with $b_{\rho,v}(u_1,u_2)$ as stated in Theorem \ref{thm:linearity}. Note also that it is immediately clear that $b_{\rho,v}(\cdot,\cdot)$ is bilinear in its arguments.

\end{proof}

\paragraph{Proof of Theorem \ref{thm:nlconv}}
%

\begin{proof}

\textbf{Case 1: $u\in C_{0}^{1}(\Omega),$ convergence in $L^{\infty}(\Omega).$}

For any $\delta>0,$ we set $A_\delta = \{t \in \mathbb{R}: |t|> \delta\},$ and we let $B_{\delta} = \mathbb{R}-A_{\delta},$ so $B_\delta = \{t \in \mathbb{R}: |t|\leq \delta\}.$
Let $u\in C_{0}^{1}(\Omega),$ and for arbitrary $\delta>0$ consider $|D_{n,v}u(x) - D_{v}u(x)|:$

\begin{align*}
|D_{n,v}u(x) - D_{v}u(x)| &= |\int_{\mathbb{R}} \frac{u(x+tv)-u(x)}{t}\rho_{n}(t)dt - D_{v}u(x)| \\
&=|\int_{\mathbb{R}} \frac{u(x+tv)-u(x) - tD_{v}u(x)}{t}\rho_{n}(t)dt|\\
&\leq \underbrace{\int_{A_{\delta}}|\frac{u(x+tv)-u(x) - tD_{v}u(x)}{t}|\rho_{n}(t)dt}_{A}\\ &+ \underbrace{\int_{B_{\delta}}|\frac{u(x+tv)-u(x) - tD_{v}u(x)}{t}|\rho_{n}(t)dt}_{B}.
\end{align*}
Consider the term $A.$ We can bound $A$ as follows:
\begin{align*}
A &\leq \frac{1}{\delta}\int_{A_\delta}|u(x+tv)-u(x)|\rho_n(t)dt + \int_{A_\delta}|D_{v}u(x)|\rho_n(t)dt\\
&\leq (\frac{2}{\delta}\|u\|_{L^{\infty}}+ |D_{v}u(x)|)\int_{A_\delta}\rho_n(t)dt\\
&\leq 2\text{ max}(\frac{2}{\delta}\|u\|_{L^{\infty}},|D_{v}u(x)|)\text{Tail}_{\delta}(\rho_n)
\end{align*}
Now we consider term $B.$

Note that $B_{\delta} = [-\delta,\delta],$ and on $B_{\delta}$ we can write 
\begin{equation*}
u(x+tv)-u(x) = tD_{v}u(\xi),
\end{equation*}
for some $\xi \in \{x+\lambda v: \lambda \in B_{\delta}\},$ since $u\in C_{0}^{1}(\Omega).$

Thus, \begin{equation*}|u(x+tv)-u(x) - tD_{v}u(x)| = |t||D_{v}u(\xi)-D_{v}u(x)|.\end{equation*}
Since $u\in C_{0}^{1}(\Omega)$, $u$ being compactly supported, it is uniformly continuous and hence given an $\epsilon$, we can find a $\mu$ such that \begin{equation*}|D_{v}u(\xi)-D_{v}u(x)|< \epsilon \, \text{ if } |\xi-x|<\mu.\end{equation*}

Since $\xi \in \{x+\lambda v: \lambda \in B_{\delta}\},$ it follows that $|\xi-x|<\delta\|v\|.$

Thus, \begin{align*}
B&=\int_{B_{\delta}}|\frac{u(x+tv)-u(x) - tD_{v}u(x)}{t}|\rho_{n}(t)dt\\
&\leq \epsilon \int_{B_{\delta}}\rho_{n}(t)dt\\
&\leq \epsilon.
\end{align*}

Finally, we can estimate

\begin{equation*}A+B \leq  2\text{ max}(\frac{2}{\delta}\|u\|_{L^{\infty}},|D_{v}u(x)|)\text{Tail}_{\delta}(\rho_n) + \epsilon.\end{equation*}
From the properties of $\rho_n$, we know that $\text{Tail}_{\delta}(\rho_n)\rightarrow 0$ as $n\rightarrow \infty$, hence 
$\overline{\text{lim}_{n}}|D_{n,v}u(x) - D_{v}u(x)| < \epsilon,$ and since $\epsilon$ was arbitrary, we conclude that $D_{n,v}u(x) \rightarrow D_{v}u(x)$ uniformly as $n\rightarrow \infty.$

\textbf{Case 2: $\text{Lip}(\Omega,M),$ convergence in $L^{\infty}(\Omega).$}

Note that the proof of Case 1 applies almost verbatim to the Lipschitz case. The only observation to be made is that if $u\in \text{Lip}(\Omega,M),$ then in particular, $u$ is uniformly continuous whose gradient (and directional derivative) exist almost everywhere with $D_{v}u \in L^{\infty}(\Omega).$ 

\textbf{Case 3: $u\in W^{1,p}(\Omega),$ convergence in $L^{p}(\Omega).$}

The general strategy for this case is using a standard density argument. Since $u\in C_{0}^{1}(\Omega)$ is dense in $W^{1,p}(\Omega),$ we first prove the convergence in $L^{p}(\Omega)$ for arbitrary $\phi \in C_{0}^{1}(\Omega)$ and then approximate $u$ in the $W^{1,p}(\Omega)$ norm by a function $\phi=\phi_u \in C_{0}^{1}(\Omega).$

Thus, first consider $\phi \in C_{0}^{1}(\Omega)$ such that $\text{supp}(\phi) \subset B(0,R)$ so that $\phi = 0 $ on the exterior $B(0,R)^C$ of the ball $B(0,R)$. Let $L>1$ be arbitrary, so that clearly $\phi = 0 $ on $B(0,LR)^C$ as well. Moreover, since $\phi = 0 $ on $B(0,LR)^C$, we have that $D_{v}\phi=0$ on $B(0,LR)^C$ too.

We now estimate as follows:

Thus, \begin{align*}
\|D_{n,v}\phi(x)-D_{v}\phi(x)\|_{L^p}&=\underbrace{\int_{B(0,LR)}|D_{n,v}\phi(x)-D_{v}\phi(x)|^p dx}_{A} \\
&+\underbrace{\int_{B(0,LR)^C}|D_{n,v}\phi(x)-D_{v}\phi(x)|^p dx}_{B}.
\end{align*}

Since $\phi \in C_{0}^{1}(\Omega),$ we know from Lemma \ref{thm:pvi} that $D_{n,v}\phi\in L^{\infty}(\Omega)$, and $D_{v}\phi \in L^{\infty}(\Omega)$ as well, being the usual directional derivative of a $C_{0}^{1}(\Omega)$ function. Thus $\|D_{n,v}\phi(x)-D_{v}\phi(x)\|_{L^{\infty}}$ is bounded as well.

We can therefore estimate term $A$ as:

\begin{align*}
A&=\int_{B(0,LR)}|D_{n,v}\phi(x)-D_{v}\phi(x)|^p dx \\
&\leq \int_{B(0,LR)}\|D_{n,v}\phi-D_{v}\phi\|^{p}_{L^{\infty}} dx \\
&\leq \|D_{n,v}\phi-D_{v}\phi\|^{p}_{L^{\infty}} \text{ vol }(B(0,LR)),
\end{align*}
where $\text{ vol }(B(0,LR))$ is the volume of the ball $B(0,LR).$ Since from Case 1, we know that $D_{n,v}\phi\rightarrow D_{v}\phi$ uniformly for $\phi \in C_{0}^{1}(\Omega),$ we conclude that given an arbitrary $\epsilon,$ there is an $N_1>0$ such that for $n>N_1,$ we have that $A<\frac{\epsilon}{2}.$

We now turn to term $B$. Recall that  $\phi(x) = 0 \text{ and } D_{v}\phi(x) = 0 $ in $B(0,LR)^C$. We have

\begin{align*}
B&=  \int_{B(0,LR)^C}|D_{n,v}\phi(x)-D_{v}\phi(x)|^p dx\\
&=\int_{B(0,LR)^C}|D_{n,v}\phi(x)|^p dx \\
&=\int_{B(0,LR)^C}|\int_{\mathbb{R}}\frac{\phi(x+tv)-\phi(x)}{t}\rho_n(t) dt|^p dx \\
&\leq_{\text{H\"older}}\int_{B(0,LR)^C}\int_{\mathbb{R}}\frac{|\phi(x+tv)-\phi(x)|^p}{|t|^p} \rho_n(t) dt dx\\
&=_{\text{Fubini}}\int_{\mathbb{R}} \int_{B(0,LR)^C}\frac{|\phi(x+tv)-\phi(x)|^p}{|t|^p} \rho_n(t) dxt dt\\
&=\int_{\mathbb{R}} \int_{B(0,LR)^C}\frac{|\phi(x+tv)|^p}{|t|^p} \rho_n(t) dx dt
\end{align*}
As in earlier proofs, we split the $\mathbb{R}-$ region of integrations into parts $A_\delta$ and $B_\delta.$ Given any $\delta>0$ set $A_\delta = \{t \in \mathbb{R}: |t|> \delta\},$ and $B_\delta = \{t \in \mathbb{R}: |t|\leq \delta\}.$
Then:
\begin{align*}
B&\leq \int_{\mathbb{R}} \int_{B(0,LR)^C}\frac{|\phi(x+tv)|^p}{|t|^p} \rho_n(t) dx dt\\
&=_{\text{Fubini}}\int_{B(0,LR)^C}\int_{\mathbb{R}} \frac{|\phi(x+tv)|^p}{|t|^p} \rho_n(t) dt dx\\
&=\int_{B(0,LR)^C}\int_{A_\delta}\frac{|\phi(x+tv)|^p}{|t|^p} \rho_n(t) dt dx\\
&+ \int_{B(0,LR)^C}\int_{B_\delta}\frac{|\phi(x+tv)|^p}{|t|^p} \rho_n(t) dt dx.
\end{align*}
Next, for $\delta < \frac{R(L-1)}{\|v\|},$ we have $\|x+tv\|>R,$ indeed, since $x\in B(0,LR)^C$ and $|t|<\delta,$ we have
\begin{align*}
\|x+tv\|&> |\|x\|-\|tv\|| = \|x\|-\|tv\|,\\
\|x\|-\|tv\|&> LR-\|v\|\frac{R(L-1)}{\|v\|} = R,
\end{align*}
so that $x+tv \in B(0,R)^C,$ and hence, in this case, $\phi(x+tv)=0.$ Thus, 
\begin{equation*}
\int_{B(0,LR)^C}\int_{A_\delta}\frac{|\phi(x+tv)|^p}{|t|^p} \rho_n(t) dt dx = 0, \, \text{ for } \delta <\frac{R(L-1)}{\|v\|}.
\end{equation*}
If $t\in B_{\delta},$ we have
\begin{align*}
&\int_{B(0,LR)^C}\int_{B_\delta}\frac{|\phi(x+tv)|^p}{|t|^p} \rho_n(t) dt dx \\
&\leq \frac{1}{\delta^p}\int_{B(0,LR)^C}\int_{B_\delta}|\phi(x+tv)|^p \rho_n(t) dt dx \\
&=_{\text{Fubini}}\frac{1}{\delta^p}\int_{B_\delta}\int_{B(0,LR)^C}|\phi(x+tv)|^p \rho_n(t)  dx dt\\
&\leq \frac{1}{\delta^p} \|\phi\|_{L^p(\Omega)}\text{Tail}_{\delta}(\rho_n) \rightarrow 0, \, n\rightarrow \infty.
\end{align*}
We therefore conclude that
\begin{align*}
\|D_{n,v}\phi(x)-D_{v}\phi(x)\|_{L^p}&=A+B \rightarrow 0, \, n\rightarrow \infty.
\end{align*}
We have thus shown for $\phi \in C_{0}^{1}(\Omega),$ that $D_{n,v}\phi(x) \rightarrow D_{v}\phi(x).$ Given now an arbitrary $u\in W^{1,p}(\Omega)$ and $\epsilon>0,$ we choose a $\phi \in C_{0}^{1}(\Omega)$ with $\|u-\phi\|_{W^{1,p}(\Omega)}<\frac{\epsilon}{3}.$ 
We have
\begin{align*}
\|D_{n,v}u-D_{v}u\|_{L^p}&\leq \underbrace{\|D_{n,v}u-D_{n,v}\phi\|_{L^p}}_{a}+\underbrace{\|D_{n,v}\phi-D_{v}\phi\|_{L^p}}_{b}+\underbrace{\|D_{v}\phi-D_{v}u\|_{L^p}}_{c}.
\end{align*}
By Lemma \ref{thm:pvi}, we have $a \leq \|v\|\|u-\phi\|_{L^p}<\|v\|\frac{\epsilon}{3}.$

Next, by what was proved earlier for $\phi,$ we have $b<\frac{\epsilon}{3}$ for some $n>N_{\epsilon}$. 

Finally $c<C(v)\frac{\epsilon}{3}$, where $C(v)$ is a constant depending on $v$. Indeed, we have $|D_{v}u(x)|\leq \|v\|\|\nabla u(x)\|,$ so that $\|D_{v}u\|_{L^p}\leq C(v) \|\nabla u(x)\|_{L^p} .$

Thus, 
\begin{align*}
\|D_{n,v}u-D_{v}u\|_{L^p}&\leq \frac{\epsilon}{3}(1+\|v\|+C(v)),
\end{align*}
and since $\epsilon$ was arbitrary, we conclude that $D_{n,v}u\rightarrow D_{v}u$ as $n\rightarrow \infty.$
\end{proof}

\paragraph{Proof of Corollary \ref{thm:nlVecconv}}
\begin{proof}
Given $x\in \mathbb{R}^N,$ we note the following equivalence of norms for $p\geq 2$:
\begin{equation*}
\|x\|_{\infty} \leq \|x\|_p \leq \|x\|_2 \leq \|x\|_1 \leq \sqrt{n}\|x\|_2\leq n\|x\|_\infty.
\end{equation*}
\paragraph{Case 1: $u\in C_{0}^{1}(\Omega)$, convergence in $L^\infty(\Omega)$:}

For $x\in \text{supp}(u)$ \begin{equation*}\|\nabla_{\overline{n},V}u(x) - V \nabla u(x)\|_{\infty} \leq \|\nabla_{\overline{n},V}u(x) - V \nabla u(x)\|_1 = \sum_{i=1}^{N}|D_{n_{i},v_{i}}u(x)- D_{v_{i}}u(x)|.\end{equation*}

As min$(\overline{n})\rightarrow \infty,$ each $n_i \rightarrow \infty,$ and from Theorem \ref{thm:nlconv}, given $\epsilon>0,$ there is an $N_\epsilon$ such that each term satisfies \begin{equation*}|D_{n_{i},v_{i}}u(x)- D_{v_{i}}u(x)|<\frac{\epsilon}{N},\end{equation*}
for min$(\overline{n})>N_{\epsilon}.$ Thus 
\begin{equation*}\sum_{i=1}^{N}|D_{n_{i},v_{i}}u(x)- D_{v_{i}}u(x)|<\epsilon,\end{equation*}
hence,
\begin{equation*}\|\nabla_{\overline{n},V}u(x) - V \nabla u(x)\|_{\infty} <\epsilon\end{equation*}
Taking the supremum over $x\in \text{supp}(u),$ we have 
\begin{equation*}\|\nabla_{\overline{n},V}u - V \nabla u\|_{L^\infty}<\epsilon,\end{equation*}
i.e.,
$\nabla_{\overline{n},V}u\rightarrow V \nabla u$ as min$(\overline{n})\rightarrow \infty.$

\paragraph{Case 2: $u\in \text{Lip}(\Omega,M)$, with compact support convergence in $L^\infty(\Omega)$ almost everywhere:}

The proof of this case is entirely similar to the previous case, after noting that $u$ is uniformly continuous whose gradient (and directional derivative) exist almost everywhere with $D_{v}u \in L^{\infty}(\Omega).$

\paragraph{Case 3: $u\in W^{1,p}(\Omega)$, convergence in $L^p(\Omega)$:}

We can write \begin{equation*}\|\nabla_{\overline{n},V}u - V \nabla u\|_{L^p}^{p} = \sum_{i=1}^{N}\|D_{n_{i},v_{i}}u- D_{v_{i}}u\|_{L^p}^{p}.\end{equation*}
From Theorem \ref{thm:nlconv}, as each $n_i \rightarrow \infty,$ we have that for any $\epsilon>0,$ there is an $N_{\epsilon}$ such that \begin{equation*}\|D_{n_{i},v_{i}}u- D_{v_{i}}u\|_{L^p}^{p}<\frac{\epsilon^p}{N},\end{equation*}
\end{proof}
for min$(\overline{n})>N_{\epsilon}.$ Thus,
\begin{equation*}\|\nabla_{\overline{n},V}u - V \nabla u\|_{L^p} < \epsilon.\end{equation*}
So that $\nabla_{\overline{n},V}u \rightarrow V \nabla u$ in $L^p(\Omega).$


\paragraph{Proof of Theorem \ref{thm:nltaylor}}
\begin{proof}
Consider $|r_{\overline{n}}(x,x_0) -r(x,x_0)|.$ 
  \begin{align*}
    |r_{\overline{n}}(x,x_0) -r(x,x_0)| &= |(x-x_0)\cdot (\nabla_{\overline{n}} \tilde{u} (x_0)-\nabla \tilde{u} (x_0))|\\
    &\leq \|x-x_0\| \| \nabla_{\overline{n}} \tilde{u} (x_0)-\nabla \tilde{u} (x_0) \| \\
    &\leq R\| \nabla_{\overline{n}} \tilde{u} (x_0)-\nabla \tilde{u} (x_0) \| \\
    &\leq RC(N)\| \nabla_{\overline{n}} \tilde{u} (x_0)-\nabla \tilde{u} (x_0) \|_{\infty},
  \end{align*}
  where $C(N)$ is a constant depending on $N$. From Corollary  \ref{thm:nlVecconv}, we have that $\| \nabla_{\overline{n}} \tilde{u} (x_0)-\nabla \tilde{u} (x_0) \|_{\infty} \leq \| \nabla_{\overline{n}} \tilde{u} -\nabla \tilde{u} \|_{L^\infty}\rightarrow 0,$ as min$(\overline{n})\rightarrow \infty$. Thus $r_{\overline{n}}(\cdot,x_0) \rightarrow r(\cdot,x_0)$ uniformly as min$(\overline{n})\rightarrow \infty$.
  
  Since $|A_{\overline{n}}(x) - A(x)| = |r_{\overline{n}}(x,x_0) -r(x,x_0)|,$ we have that $A_{\overline{n}}(x) $ converges to the standard Taylor approximant $A(x) $ as min$(\overline{n})\rightarrow \infty$.
\end{proof}


\paragraph{Proof of Theorem \ref{thm:probform}}

\begin{proof}
The proof amounts to a simple restating of the results of Lemma \ref{thm:pvi}. From Lemma \ref{thm:pvi}, we know that \begin{equation}|\mathcal{Q}^k(u,v,x,t)|\leq \|v\|^k\|u\|^{k}_{L^\infty},\end{equation} and therefore integrating with respect to $t,$ we get that 
\begin{equation}
|\mathbb{E}_{\rho_n}[\mathcal{Q}^k(u,v,x,T)]|\leq \|v\|^k\|\nabla u\|_{L^\infty}^{k}.
\end{equation}

\end{proof}

\paragraph{Proof of Theorem \ref{thm:epsilonsubgrad}}

\begin{proof}
By the convexity of $u$, we have that \begin{equation}u(y)-u(x)\geq (y-x)^{T}\nabla u(x).\end{equation}

By Corollary \ref{thm:nlVecconv}, we have that $\nabla_{\overline{n}}u(x)\rightarrow \nabla u(x)$ uniformly as min$(\overline{n})\rightarrow \infty$. For a given $\mu\in \mathbb{R}^N,$ this means, by continuity, that $\mu^T\cdot\nabla_{\overline{n}}u(x)\rightarrow \mu^T\cdot\nabla u(x).$ Choosing $\mu=(y-x),$ we see that $(y-x)^T\cdot\nabla_{\overline{n}}u(x)\rightarrow (y-x)^T\cdot\nabla u(x).$ Given an $\epsilon >0$, we can find an $N_\epsilon>0$ so that for  min$(\overline{n})>N_\epsilon,$ we have that \begin{equation}|(y-x)^T\cdot\nabla_{\overline{n}}u(x)-(y-x)^T\cdot\nabla u(x)|<\epsilon,\end{equation}which upon rearranging gives us
\begin{align}-\epsilon + (y-x)^T\cdot\nabla_{\overline{n}}u(x)&<(y-x)^T\cdot\nabla u(x)<\epsilon+(y-x)^T\cdot\nabla_{\overline{n}}u(x),\\
&\text{ and } u(y)-u(x)\geq (y-x)^{T}\nabla u(x) \\&\Rightarrow u(y)-u(x)\geq (y-x)^T\cdot\nabla_{\overline{n}}u(x) -\epsilon
\end{align}

Thus, $\nabla_{\overline{n}}u(x) \in \partial_{\epsilon}u$
\end{proof}

\paragraph{Proof of Theorem \ref{thm:nlbiased}}
Denote $b(x)= \nabla_n u(x)-\nabla u(x)$, so that we have \begin{equation}g(x,\xi) = \nabla u(x)+ b(x) + \mathcal{N}(\xi).\end{equation}
Clearly, for $\text{min}(\overline{n})>N_\epsilon,$ we have from Theorem \ref{thm:nlVecconv} that $\|b(x)\|^{2}_{L^{\infty}}<\epsilon,$ so that \begin{equation}\|b(x)\|^2 \leq \epsilon \leq \epsilon + m\|\nabla u(x)\|\end{equation}
for any $m>0.$ $\mathcal{N}(\xi)$ is bounded by definition, since 
$ \mathbb{E}[\|\mathcal{N}(\xi)\|^2]  = \sigma^2.$

Thus, both $b(x)$ and $\mathcal{N}(\xi)$ satisfy the stipulations of Theorem \ref{thm:biasedgd}, we have that that $g(x,\xi) = \nabla_n u(x) + \mathcal{N}(\xi)$ is a biased gradient.

\paragraph{Proof of Theorem \ref{thm:epsilonbiased}}
We shall need the following fact regarding $\epsilon-$subgradients from \cite{subgradoptbook}.

\begin{theorem}(Theorem 4.2.1 from \cite{subgradoptbook}) Let $u:\mathbb{R}^N\rightarrow \mathbb{R}, \, x\in \mathbb{R}^N$ and $\epsilon>0.$ Then for any $\eta>0$ and $s\in \partial_{\epsilon}u(x),$ there is an $x_\eta \in B(x,\eta)$ and $s_{\eta}\in \partial u(x_\eta)$ such that $\|s_{\eta}-s\|\leq \frac{\epsilon}{\eta}.$
\end{theorem}
This theorem imples that, roughly, an $\epsilon-$subgradient of a function $u(x)$ at $x$ can be well-approximated by a true subgradient of the same function at a nearby point $x_\eta.$ This will enable us to produce the necessary bound on $b(x).$ 
\begin{proof}
We prove the boundedness of $b(x)$ above by $\nabla u(x)$. Boundedness of $\mathcal{N}(\xi)$ is by definition, since 
\begin{equation}
\mathbb{E}[\mathcal{N}(\xi)] = 0, \, \mathbb{E}[\|\mathcal{N}(\xi)\|^2]  = \sigma^2.
\end{equation}
From the theorem stated above, we can conclude that \begin{equation}
\partial_\epsilon u(x) \subset \cap_{\eta>0}\cup_{\|y-x\|\leq \eta}\{\partial u(y) + B(0,\frac{\epsilon}{\eta})\}
\end{equation}
Thus, choosing $\eta$ appropriately, we can find an $x_\epsilon$ such that \begin{equation}x_\epsilon = x+z_\epsilon, \, \|z_\epsilon\|<\sqrt{\epsilon},\end{equation} and $b(x)\in \partial u(x_\epsilon) + B(0,\sqrt{\epsilon}).$ Since $u(x)$ is differentiable, we know $\partial u(x_\epsilon) = \{\nabla u(x_\epsilon)\}.$ Thus, we can find a $v_\epsilon \in B(0,\sqrt{\epsilon})$ such that  \begin{equation}b(x)= \nabla u(x_\epsilon) + v_\epsilon.\end{equation}

Assuming the hessian $Hu(x)$ of $u$ is uniformly bounded (which follows from smoothness of $u$), we have:

\begin{equation}\nabla u(x_\epsilon) = \nabla u(x) + Hu(x)z_\epsilon + \mathcal{O}(\epsilon),\end{equation}
so that
\begin{equation}b(x)= \nabla u(x) + Hu(x)z_\epsilon + v_\epsilon + \mathcal{O}(\epsilon),\end{equation}
and hence,
\begin{equation}\|b(x)\|^2 \leq 2\|\nabla u(x)\|^2 + 2(L^2+1)\epsilon ,\end{equation}
where $L = \|Hu\|_{L^{\infty}}.$
Thus, \begin{equation}\|b(x)\|^2 \leq m\|\nabla u(x)\|^2 + \zeta^2 ,\end{equation}
with $m=2$, and $\zeta^2 = 2(L^2+1)\epsilon.$

Since $b(x)$ and $\mathcal{N}(\xi)$ satisfy the requirements of Theorem \ref{thm:biasedgd}, we conclude that $g(x,\xi) = \nabla u(x) + b(x) + \mathcal{N}(\xi)$ is a biased gradient.

\paragraph{Remark:} The assumption that the Hessian $Hu(x)$ of $u$ is uniformly bounded is not necessary. Even a Lipschitz estimate on the gradient would suffice. Indeed, if there is an $M>0$ such that 
\begin{equation}\|\nabla u(y) - \nabla u(x)\|\leq M\|x-y\|,\end{equation}
for all $x,y$, then
\begin{equation}\|\nabla u(x_\epsilon) - \nabla u(x)\|^2\leq M^2\epsilon,\end{equation}
and hence \begin{equation}\|b(x)\|^2\leq 2\|\nabla u(x)\|^2 + 2\epsilon(M^2+1),\end{equation}
yielding a similar estimate to the one earlier.
\end{proof}

\paragraph{Proof of Theorem \ref{thm:randomdirconv}}
To ease notation, We write $dv$ to mean the measure $d\mathcal{P}_V$. We proceed in steps. 
\paragraph{Step 1: Polynomial Approximation}
Assume $u\in C_{0}^{\infty}(\Omega)$. Then, given $\epsilon>0$, $u$ can be approximated by a multivariate polynomial (by the multidimensional Stone-Weirestrass theorem, see for e.g. \citealt{peet}) $p_u$ such that

\begin{equation}
\|u-p_u(x_1,\ldots,x_N)\|_{L^{\infty}}<\frac{\epsilon}{2},
\end{equation}
for $x=(x_1,\ldots,x_N)\in\text{supp}(u),$
so that 
\begin{equation}
\|u(x+tv)-u(x)-(p_u(x+tv)-p_u(x))\|_{L^{\infty}}<\epsilon.
\end{equation}
Since $p_u(x+tv)$ is a polynomial, it can be expressed as  a convergent series of $t$:
\begin{equation}
p_u(x+tv) = p_u(x) + t\underbrace{v\cdot \nabla p_u(x)}_{L_1 p_u(x)(v)}+t^{2} \underbrace{v^{T}Hp_u(x)v}_{L_2 p_u(x)(v,v)}+\ldots,
\end{equation}
where $Hp_u(x)$ is the Hessian of $p_u$ at $x$. In general, define $V_{k}$ as follows:
\begin{equation}\label{eq:expansion}
V_k p_u (x,v) := L_k p_u(x) (\underbrace{v,v,\ldots,v}_{k-\text{factors}}),
\end{equation}
where $L_k p_u(x) (\underbrace{\cdot,\cdot,\ldots,\cdot}_{k-\text{factors}})$ is the $k-$ linear form corresponding to the (standard) Taylor expansion of $p_u(x+tv)$.

Since $p_u(x)$ is a polynomial, $L_k(\cdot)$ is continuous and $\Omega$ is bounded, we have we have that \begin{equation}\|L_k p_u(x) (\underbrace{v,v,\ldots,v}_{k-\text{factors}})\|_{L^{\infty}} \leq C_{k} O(\|v\|^{k}),\end{equation} where the constant $C_k$ depends on the operator norm of $L_k$, independent of $x\in \Omega$. Furthermore, since we assume that the distribution $\mathcal{P}_{V}$ has norm-finite moments, we have that  \begin{equation}\|\mathbb{E}_{v\sim\mathcal{P}_V}V_k p_u (x,v)\|_{L^{\infty}} < \infty \end{equation}for every $k$.

Thus, 
\begin{equation}\label{eq:expansion}
\frac{p_u(x+tv) - p_u(x)}{t} = V_1 p_u (x,v)+tV_2 p_u (x,v)+\ldots + t^{k-1}V_k p_u (x,v)+\ldots.
\end{equation}

Now,
\begin{align}\label{eq:starteqn}
\mathbb{E}_{v\sim\mathcal{P}_V}[D_{n,v}p_u(x)]&= \int_{\mathbb{R}^N}\int_{\mathbb{R}}\frac{p_u(x+tv) - p_u(x)}{t} \rho_n(t)\mathcal{P}_V(v) dt dv \\
&=\int_{\mathbb{R}^N}\int_{\mathbb{R}}(V_1 p_u (x,v)+tV_2 p_u (x,v)\\&+\ldots t^{k-1}V_k p_u (x,v)+\ldots)\rho_n(t)\mathcal{P}_V(v) dt dv \\
&=\int_{\mathbb{R}^N}v\cdot \nabla p_u(x)\mathcal{P}_V(v) dv \\&+\int_{\mathbb{R}^N}\int_{\mathbb{R}}\sum_{k\geq 2} t^{k-1} V_k p_u (x,v) \rho_n(t)\mathcal{P}_V(v) dt dv \\\label{eq:mideqn}
&=\mathbb{E}_{v\sim\mathcal{P}_V}[v]\cdot \nabla p_u(x)+\int_{\mathbb{R}^N}\int_{\mathbb{R}}\sum_{k\geq 2} t^{k-1} V_k p_u (x,v) \rho_n(t)\mathcal{P}_V(v) dt dv. 
\end{align}

We can split the terms  as follows \begin{align}\int_{\mathbb{R}^N}\int_{\mathbb{R}}\sum_{k\geq 2} t^{k-1} V_k p_u (x,v) \rho_n(t)\mathcal{P}_V(v) dt dv&=\sum_{k\geq 2}\int_{\mathbb{R}}t^{k-1} \rho_n(t)dt \int_{\mathbb{R}^N} V_k p_u (x,v)\mathcal{P}_V(v) dv.\end{align}

If we denote by $T_n$ the random variable drawn from $\rho_n(\cdot),$ we can further write:
\begin{align}\label{eq:expansion}\sum_{k\geq 2}\int_{\mathbb{R}}t^{k-1} \rho_n(t)dt \int_{\mathbb{R}^N} V_k p_u (x,v)\mathcal{P}_V(v) dv&=\sum_{k\geq 2} \mathbb{E}_{\rho_n}[T_{n}^{k-1}]\mathbb{E}_{v\sim\mathcal{P}_V}[V_{k}p_u(x,v)].\end{align}
\paragraph{Step 2: Convergence of Polynomial Moments}

We now show that for $\rho_n(\cdot) = 2^{n}\mathbb{I}_{[\frac{1}{2^{n}},\frac{1}{2^{n-1}}]}(\cdot),$ we have that $\mathbb{E}_{\rho_n}[T_{n}^{k-1}]\rightarrow 0$ as $n\rightarrow \infty$ for all $k=2,\ldots.$

For this choice of $\rho_n(\cdot),$ we have 
\begin{align}|\mathbb{E}_{\rho_n}[T_{n}^{k-1}]|\leq 2^{n}\int_{\frac{1}{2^n}}^{\frac{1}{2^{n-1}}}t^{k-1}dt &= \frac{2^n}{k}t^{k}|_{\frac{1}{2^n}}^{\frac{1}{2^{n-1}}}\\&=O(2^{-n(k-1)})O(2^k)\\&\label{eq:conv}\rightarrow 0, \text{ as }n\rightarrow \infty.\end{align}

\paragraph{Step 3: Convergence for Polynomials}
Since $p_u(x)$ is a polynomial, only finitely many terms (say up to $K-1$ terms) in expansion \ref{eq:expansion} are non-zero. Recall also that $\|\mathbb{E}_{v\sim\mathcal{P}_V}V_k p_u (x,v)\|_{L^{\infty}} < \infty$ for each $k$. Hence, we can write, using \begin{equation}\mathbb{E}_{v\sim\mathcal{P}_V}[D_{v}p_u] = \mathbb{E}_{v\sim\mathcal{P}_V}[v]\cdot \nabla p_u(x) \end{equation} and \ref{eq:starteqn} to \ref{eq:mideqn} that
\begin{align}\|\mathbb{E}_{v\sim\mathcal{P}_{V}}[D_{n,v}p_u - D_{v}p_u]\|_{L^{\infty}}&=\|\sum_{k\geq 2} \mathbb{E}_{\rho_n}[T_{n}^{k-1}]\mathbb{E}_{v\sim\mathcal{P}_V}[V_{k}p_u(x,v)]\|_{L^{\infty}}\\&\leq \sum_{k=2}^{K} |\mathbb{E}_{\rho_n}[T_{n}^{k-1}]| \|\mathbb{E}_{v\sim\mathcal{P}_V}[V_{k}p_u(x,v)]\|_{L^{\infty}}\\&\rightarrow 0 \text{ as } n\rightarrow \infty.\end{align} 

\paragraph{Step 4: Convergence for any $u\in C_{0}^{\infty}(\Omega)$}
We have so far proved for polynomials $p$ that \begin{equation}\|\mathbb{E}_{v\sim\mathcal{P}_{V}}[D_{n,v}p_u - D_{v}p_u]\|_{L^{\infty}} \rightarrow 0, \, n\rightarrow \infty.\end{equation}

To complete the proof for any $u\in C_{0}^{\infty}(\Omega),$ we choose for a given $u$ a polynomial and $\epsilon>0$ an approximant $p_u$ such that $\|u-p_u\|_L^{\infty}<\epsilon$ and $\|\nabla(u-p_u)\|_L^{\infty}<\epsilon.$ Such a choice is possible from an extended Stone-Weierstrass theorem (\citealt{peet}). We have 

\begin{equation}D_{n,v}u-D_{v}u = \underbrace{D_{n,v}u-D_{n,v}p_u}_{(1)}+\underbrace{D_{n,v}p_u-D_{v}p_u}_{(2)} + \underbrace{D_{v}p_u-D_{v}u}_{(3)}.\end{equation}

For term $(1)$ we write \begin{equation}\|D_{n,v}u-D_{n,v}p_u\|_{L^{\infty}} = \|D_{n,v}(u-p_u)\|_{L^{\infty}} \leq \|v\|\|\nabla(u-p_u)\|_{L^{\infty}} < \|v\|\epsilon,\end{equation}
where the first inequality is by Lemma \ref{thm:pvi} case 1 (continuity of $D_{n,v}(\cdot)$). Hence \begin{equation}\|\mathbb{E}_{v\sim\mathcal{P}_{V}}[D_{n,v}u-D_{n,v}p_u]\|_{L^{\infty}}<\|\mathbb{E}_{v\sim\mathcal{P}_{V}} v\|\epsilon\end{equation}

Term $(2)$ can be bounded above by $\|\mathbb{E}_{v\sim\mathcal{P}_{V}} v\|\epsilon$ for a sufficiently large $n$ from the previous discussion in Step 3. Finally, for term $(3)$ we have \begin{equation}\|\mathbb{E}_{v\sim\mathcal{P}_{V}}[D_{v}p_u-D_{v}u]\|_{L^{\infty}} \leq \|\mathbb{E}_{v\sim\mathcal{P}_{V}} v\|\epsilon, \end{equation}
since $p_u$ is an appropriate approximant to $u$.

Now, we can use the triangle inequality to combine our estimates, for a sufficiently large $n$ as 

\begin{align}\|\mathbb{E}_{v\sim\mathcal{P}_{V}}[D_{n,v}u - D_{v}u]\|_{L^{\infty}}&\leq 2\|\mathbb{E}_{v\sim\mathcal{P}_{V}} v\|\epsilon + \|\mathbb{E}_{v\sim\mathcal{P}_{V}}[D_{n,v}p_u-D_{v}p_u]\|_{L^{\infty}} \\&\leq 3\|\mathbb{E}_{v\sim\mathcal{P}_{V}} v\|\epsilon,\end{align} 
and since $\epsilon$ was arbitrary, we conclude that $\|\mathbb{E}_{v\sim\mathcal{P}_{V}}[D_{n,v}u - D_{v}u]\|_{L^{\infty}}\rightarrow 0$


\paragraph{Proof of Theorem \ref{thm:brownian}}
We first prove that $D_{n,v}W(x)$ is Gaussian.
\paragraph{$D_{n,v}W(x)$ is Gaussian:}

We write $D_{n,v}W(x)$ as:
\begin{align}
D_{n,v}W(x) &= \int \frac{W(x+tv)-W(x)}{t}\rho_n(t) dt\\
&=2^n \int_{\frac{1}{2^n}}^{\frac{1}{2^{n-1}}} \frac{W(x+tv)}{t}dt - 2^n W(x)\int_{\frac{1}{2^n}}^{\frac{1}{2^{n-1}}} \frac{1}{t}dt \\
&= 2^n \int_{\frac{1}{2^n}}^{\frac{1}{2^{n-1}}} \frac{W(x+tv)}{t}dt -2^n\text{ln}(2)W(x).
\end{align}The integral $\int_{\frac{1}{2^n}}^{\frac{1}{2^{n-1}}} \frac{W(x+tv)}{t}dt$ is expressible as the limit of a standard Riemann sum of \begin{equation}Y_N = \Delta t \sum_{k=0}^{N-1}\frac{W(x+t_kv)}{t_k},\end{equation}
\begin{equation}\int_{\frac{1}{2^n}}^{\frac{1}{2^{n-1}}} \frac{W(x+tv)}{t}dt = \lim_{N\rightarrow \infty} Y_N,\end{equation}
where $\Delta t = \frac{1}{n2^n}$ and $t_k = \frac{1}{2^n}+k\Delta t.$ Since the $N$ summands $\frac{W(x+t_kv)}{t_k}$ above are jointly multivariate normal, $Y_N$ is Gaussian, and hence the limit of $Y_N,$ namely $\int_{\frac{1}{2^n}}^{\frac{1}{2^{n-1}}} \frac{W(x+tv)}{t}dt$ is also Gaussian.
\paragraph{$D_{n,v}W(x)$ is zero mean:}

Clearly,\begin{align}
\mathbb{E}[D_{n,v}W(x)] &= \mathbb{E}[\int \frac{W(x+tv)-W(x)}{t}\rho_n(t) dt]\\
&=\int \mathbb{E}[\frac{W(x+tv)-W(x)}{t}]\rho_n(t) dt\\
&= 0,\\
\end{align}
since $\frac{W(x+tv)-W(x)}{t} \sim \mathcal{N}(0,\frac{v}{t})$ is zero mean.

\paragraph{Variance of $D_{n,v}W(x)$ = $2^{n+1}|v|(1-\text{ln}(2)):$}

We now compute the variance $\mathbb{E}[D_{n,v}W(x)]^2.$

\begin{align}
& \mathbb{E}[D_{n,v}W(x)]^2 \\=& \mathbb{E}[\int \frac{W(x+tv)-W(x)}{t} \rho_n(t) dt \int \frac{W(x+sv)-W(x)}{s} \rho_n(s) ds]\\
=&\mathbb{E}[\int\int \frac{W(x+tv)W(x+sv)-W(x)(W(x+tv)+W(x+sv))+W(x)^{2}}{ts} \rho_n(s) \rho_n(t) ds dt]\\
=&  \underbrace{\int\int \mathbb{E}[\frac{W(x+tv)W(x+sv)}{ts}]\rho_n(s) \rho_n(t) ds dt}_{(1)}
-\underbrace{\int\int \mathbb{E}[\frac{W(x)W(x+tv)}{ts}]\rho_n(s) \rho_n(t) ds dt}_{(2)}\\
-&\underbrace{\int\int \mathbb{E}[\frac{W(x)W(x+sv)}{ts}]\rho_n(s) \rho_n(t) ds dt}_{(3)} + \underbrace{\int\int \mathbb{E}[\frac{W(x)^{2}}{ts}]\rho_n(s) \rho_n(t) ds dt}_{(4)}.
\end{align}

We now compute each term $(1)-(4),$ starting with $(4)$.

\begin{align}
(4) &= \int\int \mathbb{E}[\frac{W(x)^{2}}{ts}]\rho_n(s) \rho_n(t) ds dt \\
&= x \int\int \frac{1}{ts}\rho_n(s) \rho_n(t) ds dt \\
&= x \int\frac{\rho_n(s)}{s} ds \int\frac{\rho_n(t)}{t} dt \\
&=x (2^{n})^2 [\int_{\frac{1}{2^{n}}}^{\frac{1}{2^{n-1}}} \frac{1}{\tau}d\tau]^2\\
&=x (2^{n})^2 (\text{ln}(2))^2 = (2^n\text{ln}(2))^2 x,
\end{align}
where the second equality follows from the fact that $\mathbb{E}[W(x)^2]=x,$ and we are using $\rho_{n}(\tau) = 2^{n}\mathbb{I}_{[\frac{1}{2^{n}},\frac{1}{2^{n-1}}]}.$

We next consider $(3)$. By symmetry, $(3)=(2).$ Since $0\notin \text{supp}(\rho_n(\cdot)),$ and $\text{supp}(\rho_n(\cdot)) \subset \mathbb{R}^{+},$ we have that $x+sv\neq x \, \forall s$ in the domain of integration, hence, \[ \mathbb{E}[W(x)W(x+sv)]=\min(x,x+sv)=
\begin{cases}
 x, & v\geq 0 \\
 x+sv, & v<=0. 
\end{cases}
\] 
We can therefore write for $v\geq 0:$
\begin{align}
(3) &= \int\int x \frac{1}{ts}\rho_n(s) \rho_n(t) ds dt\\
&= (2^n\text{ln}(2))^2 x,
\end{align}
while for $v<0:$
\begin{align}
(3) &= \int\int (x+sv) \frac{1}{ts}\rho_n(s) \rho_n(t) ds dt\\
&= x\int\int \frac{1}{ts}\rho_n(s) \rho_n(t) ds dt + v \int\int \frac{s}{ts}\rho_n(s) \rho_n(t) ds dt \\
&=(2^n\text{ln}(2))^2 x + v\cdot 1 \cdot 2^{n}\text{ln}(2)\\
&=2^{n}\text{ln}(2)(2^{n}\text{ln}(2)x + v),
\end{align}
so that 
\[(2)=(3)=
\begin{cases}
 (2^n\text{ln}(2))^2 x, & v\geq 0 \\
 2^{n}\text{ln}(2)(2^{n}\text{ln}(2)x + v), & v<=0. 
\end{cases}
\] 
Finally, for $(1),$ the evaluation depends on $\text{sign}(v).$
\paragraph{Case 1: $v\geq 0$}

As before, since the support of the interaction kernel $\rho_n(\cdot)$ excludes the origin, we have \begin{equation}\mathbb{E}[W(x+sv)W(x+tv)] = \min(x+tv,x+sv) = x + \min(t,s)v.\end{equation}
Hence
\begin{align}
(1) &= \int\int \mathbb{E}[\frac{W(x+sv)W(x+tv)}{ts}]\rho_n(s) \rho_n(t) ds dt \\
&= x \int\int \frac{1}{ts}\rho_n(s) \rho_n(t) ds dt + v \int\int \frac{\min(t,s)}{ts}\rho_n(s) \rho_n(t) ds dt.
\end{align}
We evaluate $\int\int \frac{\min(t,s)}{ts}\rho_n(s) \rho_n(t) ds dt$ as follows:
\begin{align}
&\int\int \frac{\min(t,s)}{ts}\rho_n(s) \rho_n(t) ds dt\\
&=\int_t \int_{s\geq t} \frac{t}{ts}\rho_n(s) \rho_n(t) ds dt +  \int_t \int_{s<t} \frac{s}{ts}\rho_n(s) \rho_n(t) ds dt\\
&=\int_t \int_{t}^{\frac{1}{2^{n-1}}} \frac{1}{s}\rho_n(s) \rho_n(t) ds dt +  \int_t \int_{\frac{1}{2^{n}}}^{t} \frac{1}{t}\rho_n(s) \rho_n(t) ds dt\\
&=\int_t 2^{n} \text{ln}(s)|_{t}^{\frac{1}{2^{n-1}}} \rho_n(t) dt +  \int_t \frac{1}{t}2^{n}(t-\frac{1}{2^{n}}) \rho_n(t) dt\\
&=2^{n} \int_t (\text{ln}(\frac{1}{2^{n-1}})-\text{ln}(t))\rho_n(t) dt +  2^{n}\int_t (1-\frac{1}{2^{n}t}) \rho_n(t) dt\\
&=2^{n} (\text{ln}(\frac{1}{2^{n-1}}) - \int_{\frac{1}{2^{n}}}^{{\frac{1}{2^{n-1}}}} \text{ln}(t)\rho_n(t) dt) +  2^{n} (1- \frac{1}{2^{n}}\int_t \frac{1}{t} \rho_n(t) dt)\\
&=_{\text{factor out } 2^{n}}2^{n} (\text{ln}(\frac{1}{2^{n-1}}) - \int_{\frac{1}{2^{n}}}^{{\frac{1}{2^{n-1}}}} \text{ln}(t)\rho_n(t) dt +   1- \frac{1}{2^{n}}\int_{\frac{1}{2^{n}}}^{{\frac{1}{2^{n-1}}}} \frac{1}{t} \rho_n(t) dt)\\
&=2^{n} (1-\text{ln}(2^{n-1}) - 2^{n}(t\text{ln}(t)-t)|_{\frac{1}{2^n}}^{\frac{1}{2^{n-1}}} - \frac{2^{n}}{2^{n}}\text{ln}(t)|_{\frac{1}{2^n}}^{\frac{1}{2^{n-1}}} )\\
&=2^{n} (1-(n-1)\text{ln}(2) - 2^{n}(\frac{1}{2^{n-1}}\text{ln}(\frac{1}{2^{n-1}})-\frac{1}{2^{n-1}} - \frac{1}{2^{n}}\text{ln}(\frac{1}{2^{n}})+\frac{1}{2^{n}})- \text{ln}(2) )\\
&=2^{n} (1-(n-1)\text{ln}(2) - \text{ln}(2) - 2^{n}(-\frac{1}{2^{n-1}}\text{ln}(2^{n-1}) + \frac{1}{2^{n}}\text{ln}(2^{n})-\frac{1}{2^{n}}) )\\
&=2^{n} (1-(n-1)\text{ln}(2) - \text{ln}(2) +2\text{ln}(2^{n-1}) -\text{ln}(2^{n})  +1 )\\
&=2^{n} (2-\text{ln}(2)((n-1)+1+n-2(n-1))) = 2^{n} (2-2\text{ln}(2))\\&= 2^{n+1} (1-\text{ln}(2)).
\end{align}
Thus, \begin{equation}(1) = (2^n \text{ln}(2))^2x+2^{n+1}v(1-\text{ln}(2)),\end{equation}
and when $v\geq 0,$ we have
\begin{align}\mathbb{E}[D_{n,v}W(x)]^2  &= (1)-(2)-(3)+(4)\\&=(2^n \text{ln}(2))^2x+v2^{n+1}(1-\text{ln}(2)) - (2^n\text{ln}(2))^2 x\\&-(2^n\text{ln}(2))^2 x+(2^n\text{ln}(2))^2 x\\&=2^{n+1}v(1-\text{ln}(2)).\end{align}
\paragraph{Case 2: $v< 0$}
In this case, we have \begin{equation}\mathbb{E}[W(x+sv)W(x+tv)] = \min(x+tv,x+sv) = x + \max(t,s)v.\end{equation}
The rest of the calculations are the mirror image of the $v\geq 0$ case, and we obtain \begin{equation}\mathbb{E}[D_{n,v}W(x)]^2 = 2^{n+1}v(\text{ln}(2)-1) = 2^{n+1}|v|(1-\text{ln}(2)).\end{equation}
Combining the two cases, we conclude that 
$\mathbb{E}[D_{n,v}W(x)]^2 = 2^{n+1}|v|(1-\text{ln}(2)).$



\vskip 0.2in
\bibliography{bibli}

\end{document}